\documentclass[letterpaper]{article} 
\usepackage{aaai25}  
\usepackage{times}  
\usepackage{helvet}  
\usepackage{courier}  
\usepackage[hyphens]{url}  
\usepackage{graphicx} 
\usepackage{multirow}
\usepackage{rotating}
\usepackage{subfigure}
\usepackage{diagbox}

\usepackage{amssymb}
\usepackage{amsfonts}
\usepackage{amsmath}
\usepackage{booktabs}
\usepackage{natbib}  
\usepackage{caption} 
\usepackage{algorithm}
\usepackage{algorithmic}
\usepackage{enumitem}
\usepackage{hyperref}

\usepackage{newfloat}
\usepackage{listings}
\DeclareCaptionStyle{ruled}{labelfont=normalfont,labelsep=colon,strut=off} 
\floatstyle{ruled}
\newfloat{listing}{tb}{lst}{}
\floatname{listing}{Listing}
\title{How to Re-enable PDE Loss for Physical Systems Modeling Under Partial Observation}
\author{
    Haodong Feng\textsuperscript{\rm 1,2} 
    Yue Wang\textsuperscript{\rm 3} \equalcontrib,
    Dixia Fan\textsuperscript{\rm 2} \equalcontrib \\
}
\affiliations{
    \textsuperscript{\rm 1} Zhejiang University 
    \textsuperscript{\rm 2} School of Engineering, Westlake University 
    \textsuperscript{\rm 3} Microsoft Research 

    \texttt{\{fenghaodong, fandixia\}@westlake.edu.cn}, \texttt{yuwang5@microsoft.com} 
}

\usepackage{bibentry}

\begin{document}

\maketitle

\begin{abstract}

In science and engineering, machine learning techniques are increasingly successful in physical systems modeling (predicting future states of physical systems).
Effectively integrating PDE loss as a constraint of system transition can improve the model's prediction by overcoming generalization issues due to data scarcity, especially when data acquisition is costly. However, in many real-world scenarios, due to sensor limitations, the data we can obtain is often only partial observation, making the calculation of PDE loss seem to be infeasible, as the PDE loss heavily relies on high-resolution states. We carefully study this problem and propose a novel framework named \textbf{\underline{R}}e-enable \textbf{\underline{P}}DE \textbf{\underline{L}}oss under \textbf{\underline{P}}artial \textbf{\underline{O}}bservation (RPLPO).
The key idea is that although enabling PDE loss to constrain system transition solely is infeasible, we can re-enable PDE loss by reconstructing the learnable high-resolution state and constraining system transition simultaneously.
Specifically, RPLPO combines an encoding module for reconstructing learnable high-resolution states with a transition module for predicting future states. The two modules are jointly trained by data and PDE loss. We conduct experiments in various physical systems to demonstrate that RPLPO has significant improvement in generalization, even when observation is sparse, irregular, noisy, and PDE is inaccurate. The code is available on \href{https://github.com/HDFengChina/RPLPO_code}{RPLPO}.

\end{abstract}

%

\section{Introduction}

Using machine learning methods to model physical systems has become a promising direction in science and engineering, e.g., fluid dynamics, diffusion, and so on \citep{an2008optimizing, zhang2023artificial, wang2023scientific}.
Since the data collection procedure is always both time and cost-consuming, a limited amount of data will hurt the generalization of the physical system modeling.
Recently, some works have introduced PDE loss in training models and achieved promising performance \citep{li2021physics, gao2021phygeonet, goswami2022physics, ren2023physr, huang2023neuralstagger} to reduce the use of costly data and improve generalization capacity of physical system models. Most works rely on high-resolution states to calculate or approximate the derivative for PDE loss. However, due to sensor limitations \citep{iakovlev2020learning, lutjens2022multiscale, yin2022continuous, boussif2022magnet, iakovlev2023learning}, the application of PDE loss for modeling physical systems is significantly challenged by the partially observable nature of measurement.
Computing partial derivatives in PDE loss with partial observation introduces significant biases, leading to a dilemma: either we avoid using PDE loss in the model, thereby weakening its generalization, or we use the biased PDE loss, which can also hurt the model's performance.
As a result, a scientific problem has been emerged:  

\textbf{\textit{Whether and how can we improve the model's generalization capacity for physical systems modeling by using PDE loss under partial observation?}}

In this work, we propose a novel framework named RPLPO to effectively re-enable PDE loss for physical systems modeling under partial observation, thereby enhancing the model's capacity to generalize and predict future partially observable states.
By using RPLPO, the PDE loss can be used to reconstruct the corresponding high-resolution state by only using partial observation and re-enable the PDE loss on the transition between adjacent learnable high-resolution states corresponding to the partial observations simultaneously.



RPLPO mainly comprises an encoding module and a transition module. The fundamental procedure is to reconstruct the learnable high-resolution state from partial observation by using the encoding module and then predict the subsequent state with the transition module. However, it is hard to train the encoding module without available high-resolution data  through data-driven supervised methods. Therefore, we use several recent consecutive observations to reconstruct a learnable high-resolution state, then train it via a PDE loss.

We design to train the encoding and transition modules jointly with two periods: a base-training period and a two-stage fine-tuning period. In base-training period, the encoding module is trained using a PDE loss without requiring high-resolution data. The transition module is trained collaboratively using the data loss and PDE loss to overcome the generalization issues due to data scarcity. In two-stage fine-tuning period, the framework leverages unlabeled data in a semi-supervised learning manner to further improve the model's generalization. The transition module is first fine-tuned with unlabeled data and PDE loss independently; then, their information is propagated to the encoding module, which is fine-tuned using data loss calculated on the original labeled data. 

We demonstrate the performance of RPLPO on five PDEs dynamics, which represent common physical systems in the real world, e.g., Burgers equation, wave equation, Navier Stokes equation, linear shallow water equation, and nonlinear shallow water equation. The results show that, with our framework, learned models have significant improvements in generalization capacity to predict future partial observed states both in single-step and multi-step. Moreover, RPLPO is validated to be: \textbf{(a)} effective in different data numbers, sparsity levels, irregular partial observations, inaccurate PDE, and noisy data; \textbf{(b)} robust in implementation and hyperparameters;  \textbf{(c)} computationally efficient under the premise of generalization improvement.

Our contributions can be summarized in two parts: \textbf{(1)} We propose a novel framework called RPLPO to re-enable PDE loss for physical systems modeling under partial observation to improve the model's generalization. \textbf{(2)} We demonstrate that RPLPO leads to a significant improvement in generalization for predicting the future partial observed state, than baseline methods, across various physical systems.

\section{Related Works} \label{related works}

We briefly discuss the highly related works here and leave the detailed version in \textbf{Technical Appendix}.

\textbf{Physics-informed machine learning:} More and more physics-informed machine learning methods have been proposed to learn the solutions of PDEs \cite{karniadakis2021physics}, including neural operator \citep{lu2019deeponet, li2020neural, li2020fourier, li2021physics, wang2021learning, gupta2021multiwavelet, goswami2022physics_1, boussif2022magnet, yin2023continuous, iakovlev2023learning, hansen2023learning, chen2022crom} and physics-informed neural networks (PINNs) \citep{raissi2019physics, yang2021b, cai2021physics, karniadakis2021physics}. These works used PDE loss to model or solve PDEs dynamics. The key to their success is the incorporation of accurate PDE, including high-resolution states or formulas. However, they cannot be applied to learn partial observation directly, as their physics-informed training manner has significant bias when calculating by partial observation. 

\textbf{High-resolution state reconstruction:} In computer vision (CV) and physical systems modeling, several works aim to reconstruct a high-resolution state from its partial observation, also known as super-resolution \cite{zhu2020beyond, li2021model}. In CV, some deep learning models \citep{dong2014learning,lim2017enhanced, soh2019natural, nazeri2019edge, zhao2020efficient}, and generative models \citep{ledig2017photo, liu2021variational, gao2023implicit} have emerged. In physical systems, the task has attracted more and more attention \citep{ren2023superbench}, which aims to use PDE loss in models \citep{wang2020physics, esmaeilzadeh2020meshfreeflownet, fathi2020super, jangid2022adaptable, ren2023physr, shu2023physics} or not need for data \citep{gao2021super,kelshaw2022physics,zayats2022super}.   Most of these works rely on high-resolution states to do supervised learning; only a small part relies on PDE loss, thus resulting in relatively larger reconstruction errors without leveraging data information or hard to model the unsteady dynamics \citep{gao2021super}. There is a work \citep{rao2023encoding} that is most related to our work. In this work, Chengpeng Rao, \textit{et al.} also considered the high-resolution state reconstruction and the time-series prediction problem, but unlike our work, it is similar to the neural PDE solvers like PINNs
rather than operator learning. It cannot generalize to different initial conditions. 
We have also leveraged this work as one of our baselines. 

Different from the above related works, our framework does not require high-resolution states but rather shows that by a carefully designed framework, the physics-informed training manner can be incorporated with partial observation and further improve model's generalization capacity.

\section{Problem Description}  \label{problem setting}
\textbf{Problem setting:} We aim to use PDE loss to improve model's generalization for predicting future partially observed states under partial observation in physical systems. 
Denote $u_t\triangleq u(t)\in\mathcal{U}$ as an observed state, we want to infer $u_{t+\tau}$ for $\tau>0$ at any time $t$. 
$\mathcal{U}$ is the functional space of form $\Omega\rightarrow\mathbb{R}^n$,
where $\Omega\subset\mathbb{R}^p$ is the set of observational point location, and $n$ is the number of system variable.
That is to say, $u_t$ is a  function of $x\in\Omega$, with vectorical output $u_t(x)\in\mathbb{R}^n$; cf. examples of Section \textbf{Experiments Setup}.
In such problems, trajectories share the same dynamics but vary by their initial conditions (ICs) $u_0\in\mathcal{U}$. We observe a finite training set of trajectories $\mathcal{D}$ with label and $\mathcal{B}$ lacks future observations (without label), using a partial spatial observation grid $\mathcal{X}\subset\Omega$ on discrete times $t\in\mathcal{T}\subset[0,T]$. In inference, the partial observation dataset is observed with $\mathcal{X}$. Note that inference is performed on test data observed from trajectories given different ICs to verify model's generalization to different trajectories.

\textbf{Evaluation scenarios:} We select four criteria that our framework should meet. First, models trained by RPLPO should be generalized to the change of trajectories to predict future partial observations. It is measured by relative error $\frac{\left \| \hat{u}_{t+\tau}-u_{t+\tau}\right \|_2}{\left \|u_{t+\tau}\right \|_2}$ as Eqn. \ref{eqn.8}. Second, it should also reconstruct accurate high-resolution states. It is measured by relative reconstruction error $\epsilon=\frac{\left \| \hat{h}_{t}-h_{t}\right \|_2}{\left \|h_{t}\right \|_2}$, where $\hat{h}_{t}, h_t$ are reconstruction and label of high-resolution states (only available to calculate $\epsilon$ in inference). Third, it should be generalized to multi-step prediction, which requires better generalization to reduce growth of errors. Finally, it should be effective for different data numbers, irregular observation, inaccurate PDE, and noisy data, robust to different encoding modules and parameters, and computationally efficient.

\section{Methodology} \label{method}
Here, we describe the overview of our framework, its key components, and our designed training strategy.
\subsection{Overview of RPLPO}\label{ref: overview}
\begin{figure*}
    \centering
    \includegraphics[width=0.9\textwidth]{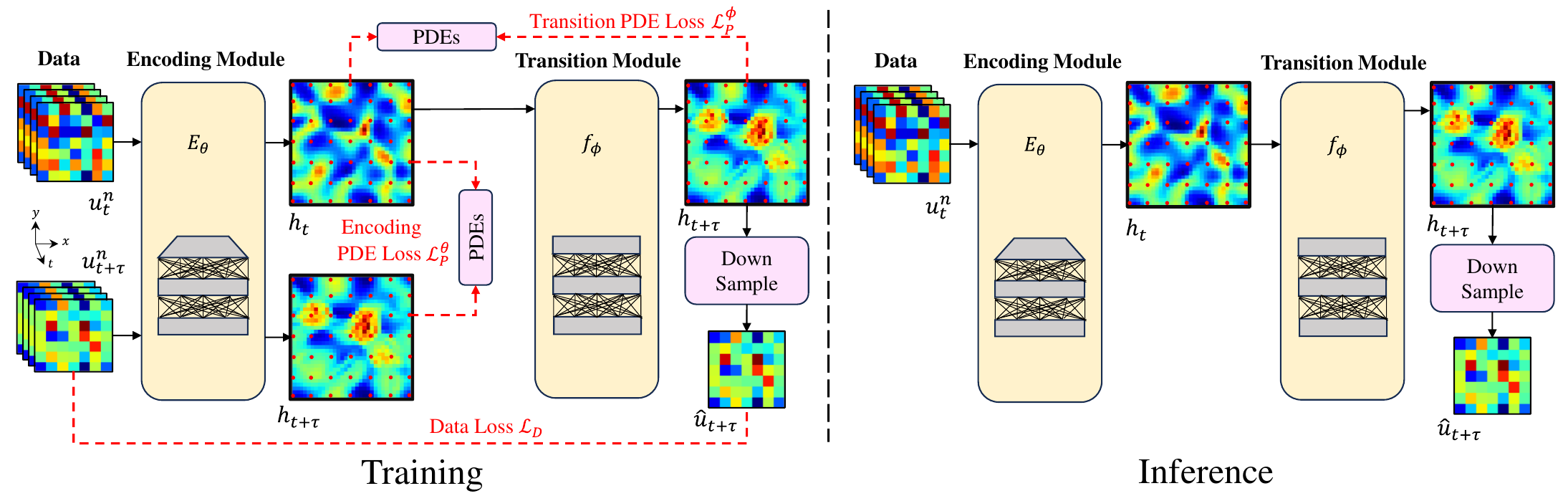}
    \caption{\textbf{RPLPO.} Training (left): In the base-training period, 
    the encoding module is trained by $\mathcal{L}_D$ and $\mathcal{L}_P^\theta$, and the transition module is trained by $\mathcal{L}_D$ and $\mathcal{L}_P^\phi$. These losses are calculated on labeled dataset $\mathcal{D}$. Then, in the two-stage fine-tuning period, the transition module is tuned by $\mathcal{L}_P^\phi$, calculated on unlabeled dataset $\mathcal{B}$, and the encoding module is tuned by $\mathcal{L}_D$, calculated on $\mathcal{D}$, in order.
    Inference (right): given the partial observation to predict future partially observed states. For the two-stage fine-tuning period, please check \textbf{Technical Appendix Figure 4}.}
    \label{fig.1}
\end{figure*}

As we mentioned in the introduction, the using of PDE loss in the model is crucial for the model generalization but is significantly challenged by partial observation.
To re-enable the use of PDE loss, RPLPO constructs PDE loss on the adjacent learnable high-resolution outputs of the encoding module. Then, the PDE loss can also facilitate the learning of transition process in the learnable high-resolution space.
RPLPO jointly trains these two modules to effectively employ PDE loss to enhance the model's generalization. As shown Fig. \ref{fig.1} (left), there are two training periods: \textbf{Base-training period:} The encoding module is trained with a PDE loss without high-resolution states, while the transition module is trained collaboratively using data loss and PDE loss. \textbf{Two-stage fine-tuning period:} It utilizes unlabeled data semi-supervisedly for further improvement. The first stage involves fine-tuning (FT) transition module independently, with PDE loss calculated on unlabeled data. Then, their information is propagated to encoding module in the second stage by fine-tuning with data loss calculated on original labeled data. 

\subsection{Model Components}\label{ref: framework}
We presented all the model components in Fig. \ref{fig.1}. Here we introduce each of them separately.

\textbf{Encoding module:} $E_\theta(u_t^n)$. The encoding module computes the learnable high-resolution state $h_t$ given the partial observation. 
To input more temporal information,
we use $n$ temporal features of $u_t^n = \{u_{t-i*\tau}\}_{i=0}^n$ to compute the more reliable $h_t$ as:
\begin{equation}\small
    h_t = E_{\theta}(u_t^n), \; t\in \mathcal{T},
\label{eqn.3}
\normalsize
\end{equation}
where $\theta$ is the trainable parameters. 

\textbf{Transition module:} $f_\phi(h_t)$. After the encoding module outputs the learnable high-resolution state $h_t$, we then model their dynamics using a neural network. Specifically, $f_\phi:\mathbb{R}^{d_h}\rightarrow\mathbb{R}^{d_h}$ to predict the subsequent high-resolution state $h_{t+\tau}$, where \(d_h\) defines the dimensionality of the high-resolution state: 
\begin{equation}\small
    h_{t+\tau}=f_\phi(h_t), \; t\in \mathcal{T},
\label{eqn.4}
\normalsize
\end{equation}
where $\phi$ denotes the trainable parameters. 

\textbf{Down-sampling:} $D(h_{t+\tau})$. 
Please note that data in $\mathcal{D}$ is partial observed, according to Fig. \ref{fig.1} (left), we define a down-sampling operation as $D: \mathbb{R}^{d_h}\rightarrow\mathcal{U}$ after the transition module, with the known coordinates. Specifically, for a learnable high-resolution state $h_{t+\tau}$, it is represented as a $n \times n$ matrix where $h_{t+\tau}(x, y)$ denotes the value at coordinates $(x, y)$.As we mentioned in \textbf{problem settings}, $\mathcal{X}\subset\Omega$ is the set of the coordinated in which we sample the partial observation. Applying the operation, the predicted partially observed state $\hat{u}_{t+\tau}$ is:
\begin{equation}\small
     \hat{u}_{t+\tau} = \{h_{t+\tau}(x, y) \mid (x, y) \in \mathcal{X}\}.
\label{eqn.6}
\normalsize
\end{equation}
By doing this, we down-sample the subsequent partially observed state $\hat{u}_{t+\tau}$ from $h_{t+\tau}$ in high-resolution space.

\textbf{Inference:} Combined altogether, our components define the inference, shown in Fig. \ref{fig.1} (right), as:
\begin{equation}\small
\begin{aligned}
     \hat{u}_{t+\tau}&=D(f_\phi(E_\theta(u_t^n))),  \; t\in \mathcal{T}.
\label{eqn.7}
\end{aligned}
\normalsize
\end{equation}

The usage of encoding module is different during training and inference periods. In training, given the ``input-output" pair ($u_t^n,u_{t+\tau}^n$), it will encode both $u_t^n$ and $u_{t+\tau}^n$ from two ends into $h_t$ and $h_{t+\tau}$ to calculate PDE loss. In inference, only input $u_t^n$ is available, and $h_t$ is computed to downstream transition module. We eliminate the uncertainty caused by missing information in partial observation by containing previous temporal observations into input, which can also be handled by Bayesian neural networks \citep{louizos2017multiplicative} and VAE \citep{kingma2013auto}, but it is not the focus of this paper. 

The ``encoding-transition-sampling" framework can leverage the PDE loss to improve the generalization of model only based on partial observation data using following our designed training strategy. 

\subsection{Model Learning}\label{ref: learning strategy}

Based on the previous components, there are three problems in training them jointly with data loss and PDE loss. The first is how to train the encoding module when we do not have the data of high-resolution states required by supervised learning. The second is how to use PDE loss to improve generalization. The third is how to use unlabeled data. To solve the above problems, we propose a learning strategy with two periods using the PDE loss function, including a base-training period given the labeled train sequences $\mathcal{D}$ and a two-stage fine-tuning period given the unlabeled data $\mathcal{B}$. We summarize the implementation in Algorithm \ref{alg.1}. 

\textbf{Loss functions:} We design the loss function with data loss and PDE loss as follows:
\begin{equation}\small
\begin{aligned}
\mathcal{L}_{PI}(\theta,\phi,\mathcal{D}) = \mathcal{L}_D(\theta,\phi,\mathcal{D})& + \gamma\mathcal{L}_{P}^\theta(\theta,\mathcal{D})+\gamma\mathcal{L}_{P}^\phi(\phi,\mathcal{D}), \\
\text{where} \; \mathcal{L}_D(\theta,\phi,\mathcal{D})&=\frac{\left \| \hat{u}_{t+\tau}-u_{t+\tau}\right \|_2}{\left \|u_{t+\tau}\right \|_2},\; \\
\mathcal{L}_P^\theta(\theta,\mathcal{D}) = F(h_t^\theta, h_{t+\tau}^\theta)^2&,\; \mathcal{L}_P^\phi(\phi,\mathcal{D}) = F(h_t^\theta, h_{t+\tau}^\phi)^2,
\label{eqn.8}
\end{aligned}
\normalsize
\end{equation} 

\begin{algorithm}[t]\small
   \caption{RPLPO}
   \label{alg.1}
    \begin{algorithmic}
   \STATE {\bfseries Input:} Dataset $\mathcal{D}$, $\mathcal{B}$, Parameters $\theta$, $\phi$, Gaps between each fine-tuning $q$, Steps of fine-tuning $m_1$, $m_2$.
   \STATE \textbf{Initialize} $\theta$ and $\phi$ of two modules $E_\theta$, $f_\phi$.
   \WHILE{True}
   \FOR{$i=1$ to $q$}
   \STATE Update $\theta$, $\phi$ using $\mathcal{L}_D$, $\mathcal{L}_{P}^\theta$, $\mathcal{L}_{P}^\phi$; for each $(u_t, u_{t+\tau})$ in $\mathcal{D}$.
   \ENDFOR
   \FOR{$i=1$ to $m_1$}
   \STATE Update $\phi$ using $\mathcal{L}_{P}^\phi$, for each $u_t$ in $\mathcal{B}$.
   \ENDFOR
   \FOR{$i=1$ to $m_2$}
   \STATE Update $\theta$ using $\mathcal{L}_D$, for each $(u_t, u_{t+\tau})$ in $\mathcal{D}$.
   \ENDFOR
   \ENDWHILE
\end{algorithmic}
\normalsize
\end{algorithm}

where $\mathcal{D}$ is the labeled dataset and $(u_t,u_{t+\tau})\in\mathcal{D}$; $\mathcal{L}_D$ is the relative $l_2$ data loss calculated on partial observation. $\mathcal{L}_P^\theta$ and $\mathcal{L}_P^\phi$ denote the PDE losses of encoding module and transition module; $\gamma$ is the weight of PDE loss; $\hat{u}_{t+\tau}$ is defined in Eqn. \ref{eqn.6}, $h_t^\theta$ and $h_{t+\tau}^\theta$ are outputs of Eqn. \ref{eqn.3}, $h_{t+\tau}^\phi$ is an output of Eqn. \ref{eqn.4}. By expressing the finite difference formulae of PDEs following \citet{huang2023neuralstagger, ren2023physr} as $F(\psi_{t},\psi_{t+\tau})=0$, where $\psi_{t}$ and $\psi_{t+\tau}$ are $t$ and $t+\tau$ solutions of PDEs, PDE losses $\mathcal{L}_P^\theta$ and $\mathcal{L}_P^\phi$ can be represented as Eqn. \ref{eqn.8}. 
We apply the widely-used standard 4th-order Runge-Kutta (RK4) and 4th-order central difference scheme in PDE loss calculation.

\textbf{Base-training period:} The base-training period can be formalized as a data loss and a PDE loss joint optimization problem that we solve in parallel: 
\begin{equation}\small
\begin{aligned}
\theta^*, \phi^* = \text{argmin}_{\theta,\phi} \; &\mathcal{L}_{PI}(\theta,\phi,\mathcal{D}).
\label{eqn.9}
\end{aligned}
\normalsize
\end{equation}
Here, we apply distinct training on each module in physics-informed manner, as shown in Fig. \ref{fig.1} (left). \textbf{(1)} For training the encoding module to offset the complete lack of fine-grained data, we propose to encode input $u_t^n$ and label $u_{t+\tau}^n$ to learnable high-resolution states $h_t$ and $h_{t+\tau}$ using the same encoding module, which can be used to calculate $\mathcal{L}_{P}^\theta$ in training. The derivative of $\mathcal{L}_{P}^\theta$ with respect to the parameter $\theta$ in the encoding module is calculated. \textbf{(2)} To train the transition module and improve generalization for predicting, $\mathcal{L}_{P}^\phi$ is computed between the input $h_{t}$ and the predicted subsequent $h_{t+\tau}$, and the derivative of $\mathcal{L}_{P}^\phi$ with respect to the parameter $\phi$ in the transition module is calculated. Along with the above physics-informed manner, the $\mathcal{L}_{D}$ is also used to train two modules end-to-end.

\textbf{Two-stage fine-tuning period:} We leverage unlabeled data to tune both modules after the base-training period to improve model's generalization further, as shown in Fig. \ref{fig.1} (left). We first optimize $\phi$ and then optimize $\theta$. The period can be formalized as an optimization problem:
\begin{equation}\small
\begin{aligned}\small
\phi^*=\text{argmin}_\phi \; \mathcal{L}_P^\phi(\phi,\mathcal{B}), \; \theta^*=\text{argmin}_\theta \; \mathcal{L}_P^\theta(\theta,\mathcal{D}),
\label{eqn.10}
\end{aligned}
\normalsize
\end{equation}
where $\mathcal{B}$ is the unlabeled dataset and the definition of $\mathcal{L}_P^\phi$ is same as Eqn. \ref{eqn.8}, except for $u_t\in\mathcal{B}$.
Please note two stages of fine-tuning have an order that is important for optimization. Intuitively, we first fine-tune the transition module with only $\mathcal{L}_{P}^\phi$ on unlabeled data $\mathcal{B}$ to make $h_{t}$ and predicted $h_{t+\tau}$ more in line with PDE loss. However, as the encoding module and transition module are first trained via $\mathcal{L}_D$ together in base-training period, the fine-tuned transition module mismatches encoding module now, leading to deteriorating performance in general. Thus, we then fine-tune encoding module with $\mathcal{L}_D$ on $\mathcal{D}$.  Because the encoding module has to be trained using both input and label, while the transition module can be trained without labels, we can propagate information of unlabeled data and PDE from transition module to encoding module by using this order. 

\section{Experiments} \label{sec5: exp}
In this section, we validate our framework's generalization capability through experiments, comparing it with baselines for both single and multi-step predictions which is more challenge due to error accumulation. 
Then, we demonstrate the effectiveness of our framework in terms of data numbers, sparsity level of observation, irregular partial observation, inaccurate PDE, and noisy data. We also assess the framework's adaptability and robustness by varying encoding modules, hyperparameters and necessity of fine-tuning. Finally, we studies the computational cost of our framework under the premise of performance by contrasting cost and GPU memory usage. Due to space limitation, we leave the details of architecture and implementation in \textbf{Technical Appendix Model Architecture and Implementation Details}.

\begin{table*}[t]
\centering
\caption{\textbf{Relative loss $\mathcal{L}_{D}$ ($\downarrow$) and $\epsilon$ ($\downarrow$) on five benchmarks.} Our framework achieves better prediction results than all other baseline methods. The results are the mean of three times running with different seeds. Best in \textbf{bold} and second best \underline{underline}.}
\label{one step result}
\renewcommand{\arraystretch}{0.85}
\setlength{\tabcolsep}{7pt}
\begin{small}
\begin{sc}
\begin{tabular}{c|c|cccccccc}
\toprule
\multicolumn{2}{c|}{Experiments} & PIDL & LNPDE & GNOT & PeRCNN & FNO & FNO* & PINO* & \textbf{RPLPO} \\
\midrule
\multirow{2}{*}{Burgers} & $\mathcal{L}_{D}$ & 1.43E-1 & 1.05E-1 & 1.93E-2 & 3.03E-1 & \underline{1.68E-2} & 1.69E-2 & 6.35E-2 & \textbf{1.37E-2} \\
 & $\epsilon$ & \underline{9.80E-6} & - & - & 0.98 & - & 6.84E-5 & 9.81E-6 & \textbf{1.85E-6} \\
\midrule
\multirow{2}{*}{Wave} & $\mathcal{L}_{D}$ & 1.28 & 9.93E-1 & 1.33E-1 & 5.44E-1 & 1.11E-1 & \underline{3.58E-2} & 1.01 & \textbf{2.64E-2} \\
 & $\epsilon$ & 1.19 & - & - & \underline{1.18} & - & 2.17 & 2.09 & \textbf{1.06} \\
\midrule
\multirow{2}{*}{NSE} & $\mathcal{L}_{D}$ & 6.45E-1 & 6.35E-1 & 1.18E-1 & 5.07E-1 & 1.06E-1 & \underline{1.68E-2} & 4.70E-1 & \textbf{1.34E-2} \\
 & $\epsilon$ & 1.71E-2 & - & - & 1.07 & - & 1.85E-2 & \underline{1.07E-2} & \textbf{2.76E-8} \\
\midrule
\multirow{2}{*}{LSWE} & $\mathcal{L}_{D}$ & 7.60 & 9.26E-2 & 1.19E-1 & 4.92E-1 & 7.92E-2 & \underline{4.75E-2} & 7.60 & \textbf{2.44E-2} \\
 & $\epsilon$ & \textbf{1.38E-3} & - & - & 0.97 & - & 6.82E-3 & 1.63E-3 & \underline{1.47E-3} \\
\midrule
\multirow{2}{*}{NSWE} & $\mathcal{L}_{D}$ & 2.34E-1 & 6.70E-2 & 1.63E-1 & 4.06E-1 & 1.01E-1 & \underline{6.41E-2} & 2.32E-1 & \textbf{3.50E-2} \\
 & $\epsilon$ & \underline{3.16E-1} & - & - & 1.30 & - & 1.74 & 3.18E-1 & \textbf{2.03E-1} \\
\bottomrule
\end{tabular}
\end{sc}
\end{small}
\end{table*}

\subsection{Experiments Setup} \label{experiment setup}

\textbf{Benchmarks:\ } We consider five PDEs that can represent common physical systems in real world. Details are in \textbf{Technical Appendix Benchmarks}.  \textbf{(1)} \textbf{Burgers equation} (Burgers) has a one-dimensional output scalar velocity field $u$.  \textbf{(2)} \textbf{Wave equation} (Wave) has the two-dimensional output velocity field $u$ and potential field $\phi$.  \textbf{(3)} \textbf{Navier Stokes equation} (NSE) corresponds to incompressible viscous fluid dynamics with three-dimensional output vector velocity ($u^x$,$u^y$) and pressure field $p$.  \textbf{(4)} \textbf{Linear shallow water equation} (LSWE) corresponds to the inviscid linearized shallow water equation with three-dimensional output vector velocity ($u^x$,$u^y$) and height $h$.  \textbf{(5)} \textbf{Nonlinear shallow water equation} (NSWE) is more challenge than LSWE, which retains nonlinear features including uneven bottom. In all benchmarks, ICs are randomly sampled from i.i.d. Gaussian Random Fields and models learn to generalize to various trajectories with different ICs, as detailed in \textbf{Technical Appendix Data Generation}.

\textbf{Baselines:\ }We reimplement several methods as baselines (details in \textbf{Technical Appendix Baseline Methods}): \textbf{(1)} \textbf{PIDL}, a physics-informed deep learning approach using finite difference-based PDE loss, previously applied in \citet{liu2021physics}. \textbf{(2)} \textbf{LNPDE} \citep{iakovlev2023learning}, a state-of-the-art model for learning physical dynamics, independent of the space-time continuous grid. \textbf{(3)} \textbf{GNOT} \citep{hao2023gnot}, an effective transformer for learning operators. \textbf{(4)} \textbf{PeRCNN} \citep{rao2023encoding}, a framework encoding physics to learn spatiotemporal dynamics. \textbf{(5)} \textbf{FNO} \citep{li2020fourier}, a neural operator with FFT-based spectral convolutions. \textbf{(6)} \textbf{FNO*}, enhancing FNO by attaching the same encoding module as in RPLPO, with $\gamma=0$ in Eqn. \ref{eqn.8}. \textbf{(7)} \textbf{PINO} \citep{li2021physics}, a hybrid approach merging data and PDE loss like FNO*, calculating PDE loss between $u_t$ and $u_{t+\tau}$. Attaching the same encoding module in FNO* and PINO* levels the playing field to evaluate our method versus modified baselines and clarify our contributions.

\subsection{Comparison With Baselines}  \label{main exp}

We compare our proposed RPLPO with seven baselines; the results are shown in Table. \ref{one step result} for single-step prediction. We visualize some prediction results in \textbf{Technical Appendix Result Visualization}. Specifically, RPLPO shows a significant improvement against all baselines on five benchmarks. RPLPO has at least 25\% (against FNO* in NSE) to more than 99\% (against PINO* in LSWE) improvement on $\mathcal{L}_{D}$ compared with all baselines. Moreover, RPLPO simultaneously has over 36\% (against PINO* in NSWE) improvement on $\epsilon$ than most of the baselines, which means RPLPO can reconstruct the more accurate high-resolution state. We consider it a reason why RPLPO can learn the physical system models with better generalization. Note that in LSWE, reconstruction error $\epsilon$ of PIDL is slightly better than ours, but its data loss $\mathcal{L}_{D}$ is much larger than ours because PIDL focuses on training the model with PDE loss only, ignoring the data constraint for predicting. In general, RPLPO achieves better performance than all baselines, which demonstrates its modeling and generalization.

\begin{figure}
    \centering
    \includegraphics[width=0.9\columnwidth]{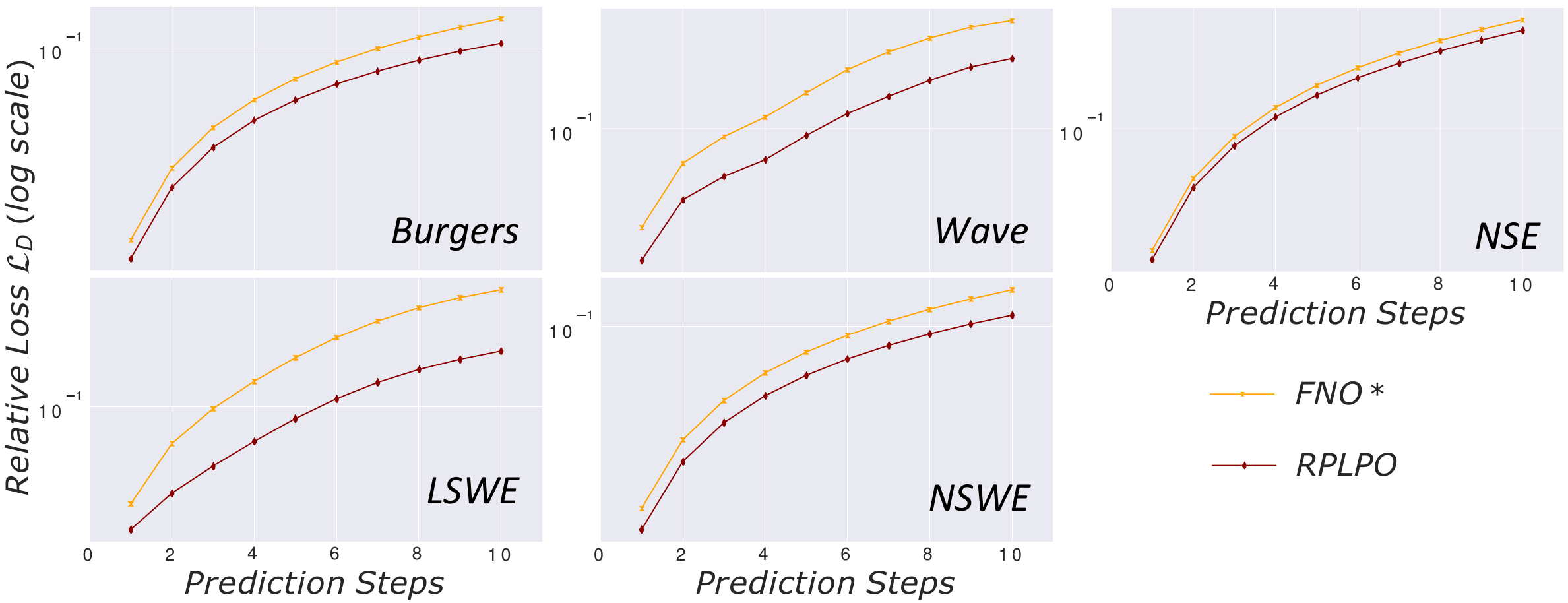}
    \caption{\textbf{The performance of multi-steps prediction from 1\textsuperscript{st} to 10\textsuperscript{th} step.} Our proposed RPLPO achieves a significant improvement against FNO* on five benchmarks among all prediction steps.}
    \label{multi step result}
\end{figure}

Furthermore,  we evaluate our framework on the multi-step prediction and compared it with baselines. The results are shown in Fig. \ref{multi step result} against FNO* (the best method among baselines), and more detailed results of others are left in \textbf{Technical Appendix Multi-step
Prediction}. Our proposed RPLPO achieves a significant improvement against FNO* on all benchmarks among prediction steps. In general, the gap between $\mathcal{L}_{D}$ of RPLPO and FNO* is increasing along with the prediction steps. The reason is that the growth of cumulative error may slow down due to the model being constrained to meet the PDE loss at each step, thus maintaining higher generalization in multi-step prediction. 

\subsection{Ablation Studies} \label{sec.abaltion}

We conduct comprehensive experiments to verify the effectiveness, robustness, and computational cost of RPLPO. The additional ablation studies, including zero-shot super-resolution, types of data loss, finite-difference schemes of PDE loss, and physics metrics, are available in the \textbf{Technical Appendix}.
\subsubsection{Effectiveness} \label{effectiveness}
In this subsection, we demonstrate how RPLPO effectively operates across data numbers, sparsity level of observation, irregular partial observation, inaccurate PDE, and noisy data. 

\textbf{Ablation study on data numbers:} We evaluate the impact of data numbers for the partial observation $\mathcal{D}$ and high-resolution states, if available, by experiments on NSWE. \textbf{(1)} Using different numbers of trajectory $|\mathcal{D}|$, and \textbf{(2)} assuming almost 1/3, 1/2, and all partial observations have corresponding high-resolution states used as labels for the encoding module, allowing more accurate state reconstruction to explore if they can improve the performance. 
\begin{table}[t]
\centering
\caption{Relative loss $\mathcal{L}_{D}$ ($\downarrow$) of the data numbers of partial observation.}
\label{data  result}
\renewcommand{\arraystretch}{1.01}
\setlength{\tabcolsep}{7pt}
\begin{small}
\begin{sc}
\begin{tabular}{c|ccc}
\toprule
$|\mathcal{D}|$ & FNO* & \textbf{RPLPO} & RPLPO w/o FT \\
\hline
50 & 1.28E-1 & \textbf{8.81E-2} & 9.53E-2 \\
100 & 1.07E-1 & \textbf{6.88E-2} & 7.34E-2 \\
300 & 6.41E-2 & \textbf{3.50E-2} & 3.69E-2 \\
350 & 5.59E-2 & \textbf{3.21E-2} & 3.32E-2 \\
\bottomrule
\end{tabular}
\end{sc}
\end{small}
\end{table}

\begin{table}[t]
\centering
\caption{Relative loss $\mathcal{L}_{D}$ ($\downarrow$) of the number of corresponding high-resolution states if available.}
\label{fine-grained available result}
\renewcommand{\arraystretch}{1.03}
\setlength{\tabcolsep}{4pt}
\begin{small}
\begin{sc}
\begin{tabular}{c|cc}
\toprule
 & FNO* & \textbf{RPLPO} \\
\hline
w/o high-resolution states & 6.41E-2 & \textbf{3.50E-2} \\
with 1/3 high-resolution states & 5.13E-2 & \textbf{3.41E-2} \\
with 1/2 high-resolution states & 5.01E-2 & \textbf{3.35E-2} \\
with all high-resolution states & 4.70E-2 & \textbf{3.29E-2} \\
\bottomrule
\end{tabular}
\end{sc}
\end{small}
\end{table}

\textbf{(i)} Table \ref{data  result} shows that $\mathcal{L}_{D}$ decreases as $|\mathcal{D}|$ increases; larger data volumes enhance model generalization. With less data, our framework still outperforms FNO*, although it's more effective with more data. \textbf{(ii)} Table \ref{fine-grained available result} reveals that both FNO* and RPLPO benefit from increased high-resolution states, with RPLPO showing greater improvements. Please note that the relative improvement of RPLPO over FNO* decreases with the increase of high-resolution states, indicating RPLPO's function in compensating for the lack of high-resolution states. 

\textbf{Sparsity level of observation and irregular observation:} 
We consider the impact of sparsity level and irregularity in partial observations. \textbf{(1)} Using different sparsity levels of observation data $x_u\times y_u$ in training. \textbf{(2)} Using irregular partial observation, sampled randomly from high-resolution states as shown in \textbf{Technical Appendix Fig. 5} to verify the performance on irregular observation. 

\textbf{(i)} Table \ref{data quality result} shows improved performance with relatively more observation ($5\times5, 7\times7$) during training, are more significant. Across various $x_u\times y_u$, RPLPO consistently outperforms FNO*.
\textbf{(ii)} For experiments on irregular observation, as shown in Table \ref{irregular result}, we compared RPLPO with two popular methods of irregular data MP-PDE \citep{brandstetter2022message} and MeshGraphNets \citep{pfaff2020learning}. RPLPO can improve performance over the baselines, consistent with the conclusion on the regular observations. 

\textbf{Inaccurate PDE and noisy data:} 
To simulate the \textbf{challenges in real-world practices}, including noisy data and inaccurate PDE, we consider the above two challenges to evaluate the performance of RPLPO. We add the unknown terms as the Gaussian random fields (GRFs) in PDE to represent the inaccuracy and use the data with different levels (10\%, 20\%, and 30\%) Gaussian noise following the previous work \cite{rao2023encoding}. We use the GRFs with mean 0 and std 1, 5, and 10. From the results in Table \ref{noise result}, we can see that RPLPO improves performance against the baseline in all settings. It illustrates that our method is effective even when the PDE is not accurate and the data has noise.
\begin{table}[t]
    \centering
        \centering
        \caption{Relative loss $\mathcal{L}_{D}$ ($\downarrow$) of different sparsity levels of partial observation.}
        \label{data quality result}
        \renewcommand{\arraystretch}{1.0}
        \setlength{\tabcolsep}{8pt}
        \begin{small}
        \begin{sc}
        \begin{tabular}{c|ccc}
        \toprule
        $x_u\times y_u$ & FNO* & \textbf{RPLPO} \\
        \hline
        $3\times3$ & 7.86E-2 & \textbf{4.43E-2} \\
        $5\times5$ & 7.08E-2 & \textbf{3.30E-2} \\
        $7\times7$ & 6.41E-2 & \textbf{3.50E-2} \\
        \bottomrule
        \end{tabular}
        \end{sc}
        \end{small}
\end{table}

\begin{table}[t]
        \centering
        \caption{Relative loss $\mathcal{L}_{D}$ ($\downarrow$) of irregular observation.}
        \label{irregular result}
        \renewcommand{\arraystretch}{1.1}
        \setlength{\tabcolsep}{4pt}
        \begin{small}
        \begin{sc}
        \begin{tabular}{cccc}
        \toprule
         MP-PDE & MeshGraphNets & FNO* & \textbf{RPLPO} \\
        \hline
         3.44E-1 & 1.85E-1 & 7.49E-2 & \textbf{5.03E-2} \\
        \bottomrule
        \end{tabular}
        \end{sc}
        \end{small}
\end{table}

\begin{table*}[t]
\centering
\caption{The increasing of computational cost (s $\downarrow$) and GPU memory utilize (MiB $\downarrow$) along the resolutions of high-resolution states increase in inference and training. Headers are three resolutions, like $64\times64$.}
\label{cost result reso}
\renewcommand{\arraystretch}{1.1}
\setlength{\tabcolsep}{4pt}
\begin{small}
\begin{sc}
\begin{tabular}{c|ccc|ccc|ccc|ccc}
\toprule
\multirow{2}{*}{Method} & \multicolumn{3}{c|}{Inference (s)} & \multicolumn{3}{c|}{Training (s)} & \multicolumn{3}{c|}{Inference (MiB)} & \multicolumn{3}{c}{Training (MiB)} \\
\cline{2-13}
 & 32  & 48 & 64 & 32 & 48 & 64 & 32  & 48 & 64 & 32 & 48 & 64 \\
\hline
FNO* & \textbf{0.0751} & 0.1445 & 0.2387 & \textbf{0.4555} & \textbf{1.5786} & \textbf{2.8061} & \textbf{3479} & \textbf{3955} & \textbf{5445} & \textbf{4311} & \textbf{5731} & \textbf{8457} \\
\textbf{RPLPO} & 0.0764 & \textbf{0.1398} & \textbf{0.2374} & 0.8007 & 2.5895 & 3.8649 & \textbf{3479} & \textbf{3955} & \textbf{5445} & 4383 & 5851 & 8615 \\
\bottomrule
\end{tabular}
\end{sc}
\end{small}
\end{table*}

\begin{table}[t]
    \centering
    \caption{Relative loss $\mathcal{L}_{D}$ ($\downarrow$) of noisy data and inaccurate PDE. The column means the noise percentages (10\%, 20\%, 30\%) on observed data. 0 means data do not have noise. The row means baseline FNO* and different scales GRFs added on the PDE used in the RPLPO. }
    \label{noise result}
    \renewcommand{\arraystretch}{1.2}
    \setlength{\tabcolsep}{4pt}
    \begin{small}
    \begin{sc}
    \begin{tabular}{c|cccc}
    \toprule
     & 0 & 10\% & 20\% & 30\% \\
    \hline
    \textbf{Accurate PDE} & 3.50E-2 & 5.05E-2 &6.11E-2 & 6.80E-2 \\
    \textbf{GRFs with std 1} & 3.52E-2 & 5.06E-2 & 6.16E-2 & 6.76E-2 \\
    \textbf{GRFs with std 5} & 4.00E-2 & 5.44E-2 & 6.30E-2 & 6.91E-2 \\
    \textbf{GRFs with std 10} & 5.12E-2 & 5.92E-2 & 6.79E-2 & 8.50E-2 \\
    FNO* & 6.41E-2 & 8.47E-2 & 8.98E-2 & 9.66E-2 \\
    \bottomrule
    \end{tabular}
    \end{sc}
    \end{small}
\end{table}


\begin{table}[t]
    \centering
    \centering
        \caption{Relative loss $\mathcal{L}_{D}$ ($\downarrow$) of RPLPO with U-Net and Transformer.}
        \vskip 0.02in
        \label{change encode result}
        \renewcommand{\arraystretch}{1.2}
        \setlength{\tabcolsep}{7pt}
        \begin{small}
        \begin{sc}
        \begin{tabular}{c|cc}
        \toprule
        Architecture & FNO* & \textbf{RPLPO} \\
        \hline
        U-Net & 6.41E-2 & \textbf{3.50E-2} \\
        Transformer & 6.79E-2 & \textbf{4.70E-2} \\
        \bottomrule
        \end{tabular}
        \end{sc}
        \end{small}
\end{table}


\subsubsection{Adaptability and Robustness}  \label{robust}

\begin{figure}[t]
\centering
\centerline{\includegraphics[scale=0.06]{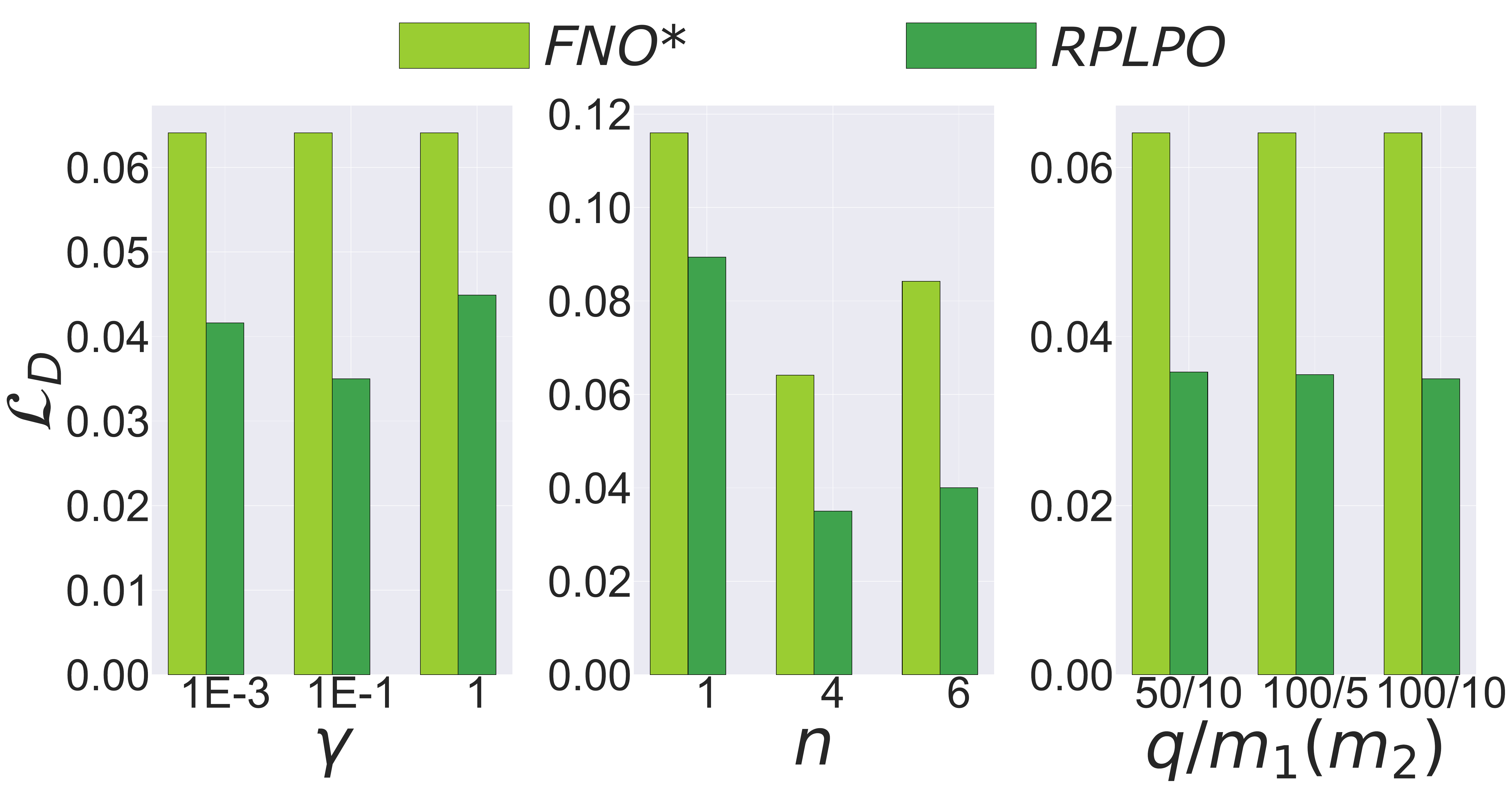}}
\caption{\small Relative loss $\mathcal{L}_{D}$($\downarrow$) of three types hyperparameters. In the right one, we omit ``$m_2$" on x-coordinate means $m_2=m_1$.\normalsize}
\label{hyper result}
\end{figure}

We change the network architecture of encoding module, hyperparameters, and RPLPO without fine-tuning to verify the framework always has improvement under different implementations. 

\textbf{Network architecture of encoding module:} To demonstrate RPLPO's independence from specific network architectures, we substitute the encoding module's U-Net with a Transformer in the NSWE setting, using ViT's official Transformer implementation \citep{dosovitskiy2020image}. Table \ref{change encode result} shows that RPLPO can still significantly outperforms the data-driven approach FNO* by using PDE loss, particularly when fine-tuning both the U-Net and Transformer encoding modules.
Moreover, we replace the PDE loss-trained encoding module with interpolation methods, including nearest neighbors (NN), bilinear, and bicubic interpolations. The result, shown in Table \ref{dcompare interpolation}, verifies the necessity and effectiveness of the proposed encoding module.


\begin{table}[H]
    \centering
    \caption{Relative reconstruction error $\epsilon$ ($\downarrow$) of the proposed encoding module and interpolations.}
    \label{dcompare interpolation}
    \renewcommand{\arraystretch}{1.05}
    \setlength{\tabcolsep}{1.5pt}
    \begin{small}
    \begin{sc}
    \begin{tabular}{c|ccccc}
    \toprule
    \small{Error} & fno* & nn & bilinear & bicubic & \textbf{RPLPO} \\
    \hline
    $\epsilon$ & 1.74 & 8.40E-1 & 5.98E-1 & 5.83E-1 & \textbf{2.03E-1} \\
    \bottomrule
    \end{tabular}
    \end{sc}
    \end{small}
\end{table}

\begin{table}[H]
    \centering
    \caption{Computational cost (s $\downarrow$) of our proposed transition module and numerical solver. Headers are three resolutions of high-resolution states, like $64\times64$.}
    \label{cost result with solver}
    \renewcommand{\arraystretch}{1.2}
    \setlength{\tabcolsep}{6pt}
    \begin{small}
    \begin{sc}
    \begin{tabular}{c|ccc}
    \toprule
    Method & $32$ (s) & $48$ (s) & $64$ (s) \\
    \hline
    \textbf{$f_\phi$ of RPLPO} & \textbf{0.0030} & \textbf{0.0032} & \textbf{0.0032} \\
    \small{ Solver} & 0.0159 & 0.0168 & 0.0175 \\
    \bottomrule
    \end{tabular}
    \end{sc}
    \end{small}
\end{table}

\textbf{Ablation study on hyperparameters:} We examine three critical hyperparameters in RPLPO for robustness: \textbf{(1)} PDE loss weight $\gamma$ from Eqn. \ref{eqn.8}, \textbf{(2)} lengths of recent temporal observations $n$ as input, and \textbf{(3)} the number of epochs $m_1, m_2$ and gaps $q$ between each fine-tuning period. \textbf{(i)} Figure \ref{hyper result} (left) indicates clearer model improvement at $\gamma$ = 1E-1, chosen for this work. RPLPO shows consistent enhancement regardless of $\gamma$, demonstrating robustness. \textbf{(ii)} Longer $n$ values ($n=4,6$) lead to better RPLPO performance, as seen in Figure \ref{hyper result} (middle), by utilizing more temporal observations. \textbf{(iii)} Figure \ref{hyper result} (right) shows RPLPO’s improvements across all parameters, confirming RPLPO can adapt to varying training setups. We also study the impact of two PDE losses with different $\gamma$ and the higher upscale factor of the encoding module. Their results are shown in \textbf{Technical Appendix Table 18 and Table 24}.

\textbf{Necessity of fine-tuning:} The two-stage fine-tuning period is an important part of RPLPO as the data scarcity tend to cause the generalization issues, especially when data acquisition is costly and only partial observation is available. We conduct the ablation study on the relationship between the data and the effect of fine-tuning.  In Table \ref{data  result}, the RPLPO w/o FT means RPLPO is not trained by the two-stage fine-tuning period. RPLPO has greater improvement than RPLPO w/o FT when data  is limited, as leveraging the unlabeled data provides additional information.

\subsubsection{Computational Cost and Analysis}

We design experiments to study the computational cost and GPU usage impacted by introducing PDE loss, and cost on the transition module on a single Nvidia V100 16GB GPU under the premise of the above effectiveness, adaptability, and robustness to demonstrate the efficiency of RPLPO. 

\textbf{Impact of PDE loss on cost and GPU usage:} We assess the rise in computational cost along the increasing of resolutions of the learnable high-resolution state. Table \ref{cost result reso} shows that regardless of resolution, FNO* and RPLPO incur similar costs and GPU memory usage due to identical model structures. However, RPLPO's training costs are higher than FNO*'s due to differing loss functions, involving calculations of PDE loss and its derivatives across two modules, but these costs remain within a reasonable range.

\textbf{Cost on transition module:} We study the computational cost of our transition module against the numerical solver (RK4) to verify the efficiency and necessity of our proposed component. As shown in Table \ref{cost result with solver}, the cost taken for a step forward using solver is more than five times that of our transition module. Moreover, as the resolution increases, the computational cost of our transition module does not significantly change, as only the input and output layers are affected, with small changes in the hidden layers. These results indicate that, in terms of computational cost, whether in training or inference, our transition module offers considerable advantages over the solver. Detail is available in \textbf{Technical Appendix Cost on Transition Module}.

\section{Conclusion}
In this paper, we propose RPLPO, a novel framework to re-enable PDE loss under partial observation to improve the model's generalization for predicting future partially observed states. Within RPLPO, we use the encoding module and transition module in order and develop a training strategy with two periods to address challenges associated with partial observation and data scarcity. 
Using five physical systems as examples, we demonstrate that RPLPO can successfully re-enable PDE loss and thus improve the model's generalization capacity. 


\section{Acknowledgements}
This work is supported by the National Natural Science Foundation of China (NSFC NO. 12302365).

\bibliography{aaai25}

\newpage
\twocolumn

\section{Technical Appendix}
\section{Model Architecture} \label{app: architecture}

Here, we detail the architecture of model with RPLPO. This architecture is used throughout all main experiments, with only adjusting a few hyperparameters (e.g., input dimension, latent dimension) depending different settings. We detail the encoding module $E_\theta$ and transition module $f_\phi$, and the architectures used in the Burgers, Wave, NSE, LSWE, and NSWE experiments. A summary of the hyperparameters is also provided in Table \ref{app hyperparameters used for structure}.

\begin{table*}[h]
\centering
\caption{Hyperparameters used for model architecture.}
\vskip 0.02in
\label{app hyperparameters used for structure}
\renewcommand{\arraystretch}{1.2}
\setlength{\tabcolsep}{4pt}
\begin{small}
\begin{sc}
\begin{tabular}{l|ccccc}
\toprule
Hyperparameters for model & Burgers & Wave & NSE & LSWE & NSWE \\
\hline
$E_\theta$: Input dimension  & $(n,7,7)$ & $(2n,9,9)$ & $(3n,7,7)$ & $(3n,7,7)$ & $(3n,7,7)$ \\
$E_\theta$: Output dimension  & $(1,32,32)$ & $(2,41,41)$ & $(3,32,32)$ & $(3,32,32)$ & $(3,32,32)$ \\
$E_\theta$: Residual block number & 23 & 23 & 23 & 23 & 23 \\
$E_\theta$: Channel & 32 & 32 & 32 & 32 & 32 \\
$E_\theta$: Dropout & 0.1 & 0.1 & 0.1 & 0.1 & 0.1 \\
$E_\theta$: Activation function & Swish & Swish & Swish & Swish & Swish \\
$f_\phi$: Input dimension & $(1,32,32)$ & $(2,41,41)$ & $(3,32,32)$ & $(3,32,32)$ & $(3,32,32)$ \\
$f_\phi$: Output dimension & $(1,32,32)$ & $(2,41,41)$ & $(3,32,32)$ & $(3,32,32)$ & $(3,32,32)$ \\
$f_\phi$: Layers number & 4 & 4 & 4 & 4 & 4 \\
$f_\phi$: Modes & 12 & 12 & 12 & 12 & 12 \\
$f_\phi$: Width & 32 & 32 & 32 & 32 & 32 \\
$f_\phi$: Activation function & GeLu & GeLu & GeLu & GeLu & GeLu \\
Down-sample output & $(1,7,7)$ & $(2,9,9)$ & $(3,7,7)$ & $(3,7,7)$ & $(3,7,7)$ \\
\bottomrule
\end{tabular}
\end{sc}
\end{small}
\end{table*}

\textbf{Burgers}: For the experiments in Burgers equation, our encoding module $E_\theta$ employs a U-Net with the number of 23 residual blocks. The input dimension is $(n,7,7)$, and the output dimension of the encoding module is $(1,32,32)$, which is the high-resolution state. The transition module $f_\phi$ employs an FNO model with the number of 4 FNO layers and a layer width of 32. It uses the GeLu activation \citep{li2020fourier}. The input dimension is $(1,32,32)$, and that of the output is $(1,32,32)$. The down-sampling block employs a gap of 5 to get the predicted subsequent partially observed state that has $(1,7,7)$ dimension. For both partial observation and high-resolution state, one channels of the first dimension represent the $u$ that is the field of velocity in Eqn. \ref{eqn.burger}. The second and third dimensions are the spatial dimensions, which are 7 for partial observation and 32 for high-resolution state. In this setting, the observed proportion is less than 4.79\%.

\textbf{Wave}: For the experiments in wave equation, our encoding module $E_\theta$ employs a U-Net with the number of 23 residual blocks. The input dimension is $(2n,9,9)$, and the output dimension of the encoding module is $(2,41,41)$, which is the high-resolution state. The transition module $f_\phi$ employs an FNO model with the number of 4 FNO layers and a layer width of 32. It uses the GeLu activation \citep{li2020fourier}. The input dimension is $(2,41,41)$, and that of the output is $(2,41,41)$. The down-sampling block employs a gap of 5 to get the predicted subsequent partially observed state that has $(2,9,9)$ dimension. For both partial observation and high-resolution state, two channels of the first dimension represent the $u$ that is the field of velocity in Eqn. \ref{eqn.wave}, and $\phi$ that is the field of velocity potential quite related to $u$ in wave equation. The second and third dimensions are the spatial dimensions, which are 9 for partial observation and 41 for high-resolution state. In this setting, the observed proportion is less than 4.82\%.

\textbf{NSE}: For the experiments in shallow water equation, our encoding module $E_\theta$ employs a U-Net with the number of 23 residual blocks. The input dimension is $(3n,7,7)$, and the output dimension of the encoding module is $(3,32,32)$, which is the high-resolution state. The transition module $f_\phi$ employs an FNO model with a number of 4 FNO layers, and the width is 32. It uses the GeLu activation \citep{li2020fourier}. The input dimension is $(3,32,32)$, and that of the output is $(3,32,32)$. The down-sampling block employs a gap of 5 to get the predicted subsequent partially observed state that has $(3,7,7)$ dimension. For both partial observation and high-resolution state, three channels of the first dimension represent the $u^x$ and $u^y$ that are the fields of velocity in $x$ and $y$ directions, and $p$ that is the field of pressure, in Eqn. \ref{eqn.nse}. The second and third dimensions are the spatial dimensions, which are 7 for partial observation and 32 for high-resolution state. In this setting, the observed proportion is less than 4.79\%.

\textbf{LSWE and NSWE}: For the experiments in shallow water equation, our encoding module $E_\theta$ employs a U-Net with the number of 23 residual blocks. The input dimension is $(3n,7,7)$, and the output dimension of the encoding module is $(3,32,32)$, which is the high-resolution state. The transition module $f_\phi$ employs an FNO model with a number of 4 FNO layers, and the width is 32. It uses the GeLu activation \citep{li2020fourier}. The input dimension is $(3,32,32)$, and that of the output is $(3,32,32)$. The down-sampling block employs a gap of 5 to get the predicted subsequent partially observed state that has $(3,7,7)$ dimension. For both partial observation and high-resolution state, three channels of the first dimension represent the $u^x$ and $u^y$ that are the fields of velocity in $x$ and $y$ directions, and $h$ that is the field of fluid column height, in Eqn. \ref{eqn.lswe} and \ref{eqn.nswe}. The second and third dimensions are the spatial dimensions, which are 7 for partial observation and 32 for high-resolution state. In this setting, the observed proportion is less than 4.79\%.

\section{Implementation Details}\label{ref: detail of implementation}

In this section, we provide experiment details for five benchmarks. First, we introduce the details of data generation. Then, we introduce the manner of PDE loss calculation during the training. After that, the details about base-training and two-stage fine-tuning periods are supplemented. Table \ref{app hyperparameters used for training} shows general hyperparameters of training, except for the hyperparameters that are modified in ablation studies. The models are trained and evaluated on a single Nvidia V100 16GB GPU. Moreover, we detail the experiments setup and ablation studies.

\begin{table}[h]
\centering
\caption{Hyperparameters used for training.}
\vskip 0.02in
\label{app hyperparameters used for training}
\renewcommand{\arraystretch}{1.2}
\setlength{\tabcolsep}{4pt}
\begin{small}
\begin{sc}
\begin{tabular}{l|c}
\toprule
Hyperparameters name for training & Value \\
\hline
Batch size & 32  \\
$\alpha$, Learning rate & 1 \\
$\lambda$: Weight decay & 1E-4 \\
$\delta t$: Time gap & 0.01 \\
$\gamma$: Scheduler factor & 0.5 \\
Training epochs & 1000 \\
\bottomrule
\end{tabular}
\end{sc}
\end{small}
\end{table}

\subsection{Data Generation}\label{app: detail of data generation}
To generate the training data, we initiate initial conditions (ICs) randomly sample from the i.i.d. GRFs and subsequently evolved them both spatially and temporally. Specifically, we employed the RK4 method for temporal evolution, starting from $t = 0$ and progressing up to $t = 1$ with a time-step of $\delta t = 0.01$ in the same way to \citet{rosofsky2023applications}. For the computation of spatial derivatives, a fourth-order central difference scheme was utilized within the framework of FDM. As we aim to generalize the model on a variety of trajectories, we divide the training dataset $\mathcal{D}$, test dataset, and unlabeled dataset $\mathcal{B}$ based on trajectories generated from different ICs. 

\subsection{Calculation of PDE loss}\label{ref:detail of PDE loss}
PDE loss both $\mathcal{L}_{P}^{\theta}$ and $\mathcal{L}_{P}^{\phi}$ are calculated as similar method like data generation. By discretizing the PDEs with fourth-order central-difference scheme and the RK4 time discretization method on high-resolution state ($41\times 41$ in Wave, $32\times32$ in others), the RK4 to solve differential equation $\psi'=P(x,y,\psi)$ can be expressed as:
\begin{equation}
\begin{aligned}
k_{1} &= P(x, y, \psi_t), \\
k_{2} &= P\left(x + \frac{w}{2}, y + \frac{w}{2}, \psi_t + \frac{w}{2}k_{1}\right), \\
k_{3} &= P\left(x + \frac{w}{2}, y + \frac{w}{2}, \psi_t + \frac{w}{2}k_{2}\right), \\
k_{4} &= P\left(x + w, y + w, \psi_t + wk_{3}\right), \\
\psi_{t+\tau} &= \psi_t + \frac{w}{6}(k_{1} + 2k_{2} + 2k_{3} + k_{4}).
\end{aligned}
\end{equation}
Let a function $F$ denotes the RK4 calculation, where $F(\psi_t,\psi_{t+\tau})=0$. We design $F^2$ as the PDE loss. Specifically, to the encoding PDE loss:
\begin{equation}
\begin{aligned}
\mathcal{L}_{P}^{\theta}(\theta,\mathcal{D}) = F(h_t^{\theta}, h_{t+\tau}^{\theta})^2.
\end{aligned}
\end{equation}
To the transition PDE loss:
\begin{equation}
\begin{aligned}
\mathcal{L}_{P}^{\phi}(\phi,\mathcal{D}) = F(h_t^{\theta}, h_{t+\tau}^{\phi})^2.
\end{aligned}
\end{equation}

\begin{algorithm}[tb]\small
   \caption{RPLPO (Detailed Version)}
   \label{alg.2}
\begin{algorithmic}
   \STATE {\bfseries Input:} Labeled dataset $\mathcal{D}$ and unlabeled dataset $\mathcal{B}$, Parameters $\theta$ and $\phi$, Gaps between each fine-tuning periods $q$, Steps of two-stage fine-tuning $m_1$ and $m_2$.
   \STATE \textbf{Initialize} parameters $\theta$ of the encoding module $E_\theta$, and $\phi$ of the transition module $f_\phi$.
   \WHILE{True}
   \FOR{$i=1$ to $q$}
   \FOR{each $(u_t, u_{t+\tau})$ in $\mathcal{D}$}
   \STATE Encode $u_t^n$ to $h_t$ using $E_{\theta}$, encode $u_{t+\tau}^n$ to $h_{t+\tau}$ using $E_{\theta}$.
   \STATE Calculated $\mathcal{L}_{P}^\theta$.
   \STATE Predict $h_{t+\tau}$ using $f_{\phi}$.
   \STATE Calculate $\mathcal{L}_{P}^\phi$.
   \STATE Down-sample $u_{t+\tau}$.
   \STATE Calculate $\mathcal{L}_D$.
   \STATE Update $\theta$, $\phi$ using $\mathcal{L}_D$, $\mathcal{L}_{P}^\theta$, $\mathcal{L}_{P}^\phi$.
   \ENDFOR
   \ENDFOR
   \FOR{$i=1$ to $m_1$}
   \FOR{each $u_t$ in $\mathcal{B}$}
   \STATE Encode $u_t^n$ to $h_t$ using $E_{\theta}$.
   \STATE Predict $h_{t+\tau}$ using $f_{\phi}$.
   \STATE Calculate $\mathcal{L}_{P}^\phi$.
   \STATE Update $\phi$ using $\mathcal{L}_{P}^\phi$.
   \ENDFOR
   \ENDFOR
   \FOR{$i=1$ to $m_2$}
   \FOR{each $(u_t, u_{t+\tau})$ in $\mathcal{D}$}
   \STATE Encode $u_t^n$ to $h_t$ using $E_{\theta}$.
   \STATE Predict $h_{t+\tau}$ using $f_{\phi}$.
   \STATE Down-sample $u_{t+\tau}$.
   \STATE Calculate $\mathcal{L}_D$.
   \STATE Update $\theta$ using $\mathcal{L}_D$.
   \ENDFOR
   \ENDFOR
   \ENDWHILE
\end{algorithmic}
\normalsize
\end{algorithm}

\subsection{Model Learning}\label{ref:detail of learning strategy}

\textbf{Base-Training Period}: In the base-training period, we jointly train both the encoding module and the transition module. This is done using a combination of data-driven and physics-informed methods. A significant challenge we face is the absence of high-resolution data, as we can only access the partial observation $u_t$ and $u_{t+\tau}$. To solve this challenge, we introduce PDE loss terms $\mathcal{L}_{P}^{\theta}$ and $\mathcal{L}_{P}^{\phi}$ for the encoding and transition modules, respectively. The encoding module is trained to map the partial observed inputs $u_t$ and $u_{t+\tau}$ to learnable high-resolution states $h_t$ and $h_{t+\tau}$, which can then be used to compute the $\mathcal{L}_{P}^{\theta}$. In a similar way, the transition module is trained using $\mathcal{L}_{P}^{\phi}$. By collaboratively training these modules via $\mathcal{L}_D$, $\mathcal{L}_{P}^{\theta}$ and $\mathcal{L}_{P}^{\phi}$, we are able to reconstruct the more reliable high-resolution state without requiring high-resolution data and improve the generalization capacity of physical system models, thereby overcoming the limitations of data-driven supervised approaches. Please note that in this period, we train the model not only using PDE losses respectively but also using the data loss end-to-end to avoid the inappropriate application of PDE loss. To facilitate the training of the encoder module using PDE loss, we replace the corresponding pixels of the learnable high-resolution state with the observed pixels and apply the stop-gradient operation on them.

\begin{figure*}[tb]
\centering
\centerline{\includegraphics[width=0.9\textwidth]{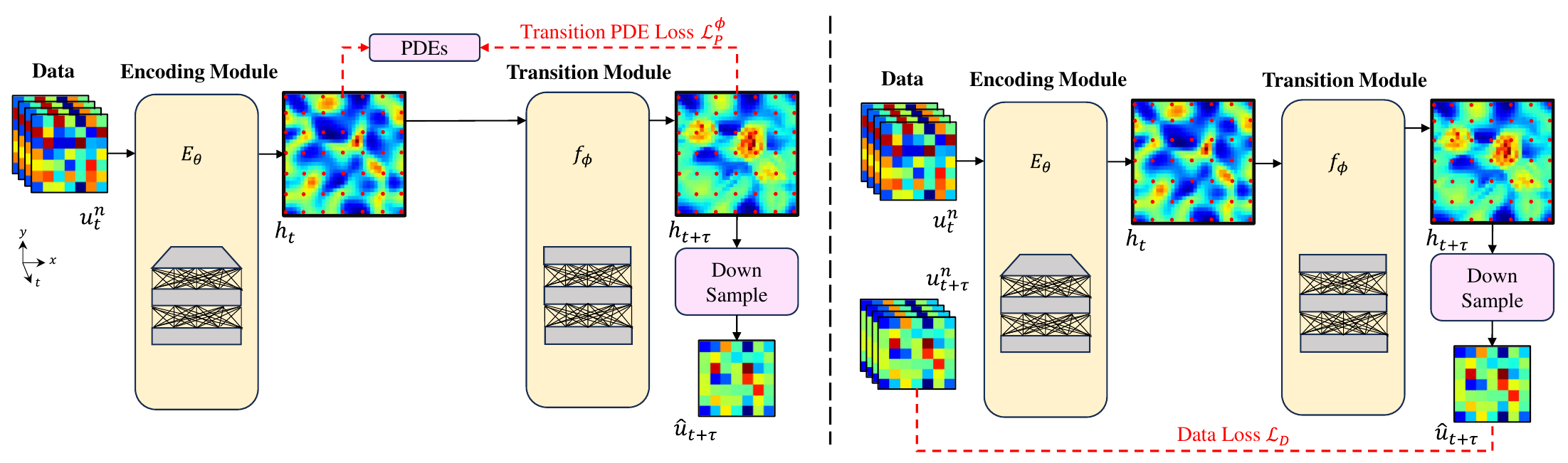}}
\caption{\textbf{Two-Stage Fine-Tuning Period.} The first stage (left) and the second stage (right).}
\label{fig.6}  
\end{figure*}

\textbf{Two-Stage Fine-Tuning Period}: In two-stage fine-tuning period, given the abundance of readily available unlabeled, partial observation for PDE-governed physical systems, we first focus on the transition module in the first stage. In this stage, the module is fine-tuned independently using the $\mathcal{L}_{P}^{\phi}$ calculated using the unlabeled input (learnable high-resolution state) and the predicted subsequent high-resolution state, thereby enhancing its prediction to better align with PDE. However, this tuned transition module is not inherently compatible with the encoding module, which was trained during the base-training period. This incongruence results in a gradual deterioration on performance. To address this problem, a second fine-tuning stage is introduced. Here, the encoding module is fine-tuned individually using the data loss $\mathcal{L}_{D}$, which is computed based on the original training set $\mathcal{D}$ without the introduction of new labeled data. By proceeding in this manner, we effectively propagate the information of PDE and unlabeled data from the transition module into the encoding module. The RPLPOine of fine-tuning is illustrated in Fig. \ref{fig.6}. The detailed implementation summary is in Algorithm \ref{alg.2}.

\subsection{Experiments Setup}

\subsubsection{Benchmarks} \label{app benchmarks}
We provide additional details on five benchmarks used in experiments.

$\bullet$ \textbf{Burgers equation} (Burgers): It has has a one-dimensional output scalar velocity field $u$.
\begin{equation}\small
\begin{aligned}
\frac{\partial u}{\partial t}+\frac{\partial (u^2/2)}{\partial x}+\frac{\partial (u^2/2)}{\partial y} = \nu(& \frac{\partial^2 u}{\partial x^2}+\frac{\partial^2 u}{\partial y^2}),
\label{eqn.burger}
\end{aligned}
\normalsize
\end{equation}
where $x,y \in [0,1), t \in[0,1]$, $u(x,y,0) =u_0(x,y)$, $u$ is velocity field, $\nu$ denotes the viscosity coefficient. 

$\bullet$ \textbf{Wave equation} (Wave): It has the two-dimensional output velocity field $u$ and potential field $\phi$.
\begin{equation}\small
\begin{aligned}
\frac{1}{{c}^{2}}\frac{\partial^2\phi}{\partial x^2}&=\nabla^2\phi,\\
u&=-\nabla \phi,
\label{eqn.wave}
\end{aligned}
\normalsize
\end{equation}
where $x,y \in [0,1), t \in[0,1]$, $u(x,y,0) =u_0(x,y)$, $u$ is the velocity field, $\phi$ is the field of velocity potential, $c$ denotes the speed of wave. 

$\bullet$ \textbf{Navier Stokes equation} (NSE): It corresponds to the incompressible viscous fluid dynamics with three-dimensional output vector velocity fields ($u$,$v$) and pressure field $p$.
\begin{equation}\small
\begin{aligned}
\frac{\partial u^x}{\partial t} + u^x \frac{\partial u^x}{\partial x} + u^y \frac{\partial u^x}{\partial y} &= -\frac{1}{\rho} \frac{\partial p}{\partial x} + \nu \left( \frac{\partial^2 u^x}{\partial x^2} + \frac{\partial^2 u^x}{\partial y^2} \right) + f_x, \\
\frac{\partial u^y}{\partial t} + u^x \frac{\partial u^y}{\partial x} + u^y \frac{\partial u^y}{\partial y} &= -\frac{1}{\rho} \frac{\partial p}{\partial y} + \nu \left( \frac{\partial^2 u^y}{\partial x^2} + \frac{\partial^2 u^y}{\partial y^2} \right) + f_y, \\
\frac{\partial u^x}{\partial x} + &\frac{\partial u^y}{\partial y} = 0,
\label{eqn.nse}
\end{aligned}
\normalsize
\end{equation}
where $x,y\in [0,1)$, $t\in [0,1]$, $p(x,y,0)=p_{0}(x,y)$, $u^x(x,y,0)=0$, $u^y(x,y,0)=0$, $p$ is the pressure field, $u^x$ and $u^y$ are the velocity fields of $x$ and $y$ directions. $\nu$ denotes the viscosity coefficient. $f_x$ and $f_y$ are the body forces per unit mass in the $x$ and $y$ directions. The last equation represents the continuity equation for incompressible flow.

$\bullet$ \textbf{Linear shallow water equation} (LSWE): It corresponds to the inviscid linearized shallow water equation with three-dimensional output vector velocity fields ($u$,$v$) and height $h$.
\begin{equation}\small
\begin{aligned} 
\frac{\partial h}{\partial t}+H(\frac{\partial u^x}{\partial x} & +\frac{\partial u^y}{\partial y})=0, \\
\frac{\partial u^x}{\partial t}-f u^y =-g \frac{\partial h}{\partial x}, \quad & \frac{\partial u^y}{\partial t}+f u^x =-g \frac{\partial h}{\partial y},
\label{eqn.lswe}
\end{aligned}
\normalsize
\end{equation}
where $x,y\in [0,1)$, $t\in [0,1]$, $h(x,y,0)=h_{0}(x,y)$, $u^x(x,y,0)=0$, $u^y(x,y,0)=0$, $h$ is the surface height, $u^x$ and $u^y$ are the velocity fields of $x$ and $y$ directions, $H$ is the mean height, $f$ is the Coriolis coefficient, and $g$ is the gravitational constant.

$\bullet$ \textbf{Nonlinear shallow water equation} (NSWE): The formula of NSWE is Eqn. \ref{eqn.nswe}. It is a more challenging benchmark with some adjustments based on that in \citet{rosofsky2023applications}, which is more useful in the applications, as NSWE retains all nonlinear terms, including the uneven bottom. It has three-dimensional output vector velocity fields ($u$,$v$) and height $h$. In Eqn. \ref{eqn.nswe}, $x,y\in[0,1)$, $t\in[0,1]$, $h(x,y,0)=h_{0}(x,y)$, $u^x(x,y,0)=0$, $u^y(x,y,0)=0$, $z$ is the uneven height of the bottom randomly sampled from GRFs, $\nu$ denotes the viscosity coefficient. 

\begin{figure*}[t]
\begin{equation}\small
\begin{aligned}
\frac{\partial(h)}{\partial t}+\frac{\partial(h u^x)}{\partial x} & +\frac{\partial(h u^y)}{\partial y} =0, \\
\frac{\partial(h u^x)}{\partial t} +\frac{\partial}{\partial x}[h (u^x)^2+\frac{1}{2} g h^2]+\frac{\partial(h u^x u^y)}{\partial y} & =\nu\left(\frac{\partial^2 u^x}{\partial x^2}+\frac{\partial^2 u^x}{\partial y^2}\right) - fu^y + gh\frac{\partial z}{\partial x}, \\
\frac{\partial(h u^y)}{\partial t}+\frac{\partial(h u^x u^y)}{\partial x}+\frac{\partial}{\partial y}[h (u^y)^2+\frac{1}{2} g h^2] & =\nu\left(\frac{\partial^2 u^y}{\partial x^2}+\frac{\partial^2 u^y}{\partial y^2}\right) + fu^x + gh\frac{\partial z}{\partial y}.
\label{eqn.nswe}
\end{aligned}
\normalsize
\end{equation}
\end{figure*}

\subsubsection{Baseline Methods}  \label{app baselines}
Here, we provide additional details on the baselines used in experiments.

$\bullet$ \textbf{PIDL}: A physics-informed deep learning methodology that is purely based on soft PDE constraints computed by finite difference and do not use training data which is utilized in a model-based reinforcement learning \citep{liu2021physics}. As we aim to compare the performance between RPLPO and the model purely trained by PDE loss, PIDL has the same architecture as RPLPO except for training via PDE loss independently calculated as \citet{liu2021physics} in this paper.

$\bullet$ \textbf{LNPDE}: It is a space-time continuous grid-independent model for learning physical dynamics from partial observations \citep{iakovlev2023learning}. We employ the official implementation and compare with our proposed RPLPO.

$\bullet$ \textbf{FNO}: It is a powerful neural operator with FFT-based spectral convolutions \citep{li2020fourier}. To make FNO have learned parameters that are of comparable scale as RPLPO, we employ FNO with 5 layers, and the width is 48.

$\bullet$ \textbf{FNO*}: Based on FNO, we modify it to a similar architecture like RPLPO, which attaches an encoding module before the FNO model. So that, we can create a more level playing field for evaluating the effectiveness of our proposed method against the modified baseline, thereby providing a clearer measure of our contributions.

$\bullet$ \textbf{PINO*}: PINO \citep{li2021physics} is a hybrid approach incorporating data and PDE loss based on FNO to learn the neural operator. We also introduced a specific modification, like what we do in FNO*, to better address the nuances of our research problem, called PINO*. The input and output are in partial observed mesh, which is used to calculate the PDE loss. By using this baseline, we compare which is better to calculate PDE loss on learnable high-resolution state or partial observation.

$\bullet$ \textbf{GNOT}: GNOT \citep{hao2023gnot} is a general neural operator transformer (GNOT), a scalable and effective transformer-based framework for learning operators.  We employ the official implementation and compare with our proposed RPLPO.

$\bullet$ \textbf{PeRCNN}: PeRCNN \citep{rao2023encoding} is a deep learning framework that forcibly encodes given physics structure to facilitate the learning of the spatiotemporal dynamics in sparse data regimes. As the main problem studied in this paper is similar to the work of neural PDE solvers like PINN, rather than operator learning, we modified the model to the operator learning setting like ours that only uses the high-resolution data, and generalizes to the different initial conditions and trajectories. 

\subsection{Comparison with CROM}
We conducted experiments on a related baseline CROM \cite{chen2022crom} in the NSWE setting (the most challenge benchmark in the paper) following the official code. From the results in the following table, we can see that our method achieves better performance.

\begin{table}[H]
    \centering
    \caption{Relative loss of RPLPO and CROM}
    \label{inaccurate nu}
    \renewcommand{\arraystretch}{1.2}
    \setlength{\tabcolsep}{4pt}
    \begin{small}
    \begin{sc}
    \begin{tabular}{c|cc}
    \toprule
    $\nu$ & \textbf{RPLPO} & CROM  \\
    \hline
     $\mathcal{L}_{D}$ & 3.50E-2 & 6.75E-2 \\
    \bottomrule
    \end{tabular}
    \end{sc}
    \end{small}
\end{table}

\subsection{Details of Ablation Studies} \label{app_ablation}

\subsubsection{Ablation Study on Data Numbers}
In this part, we first evaluate the impact of the  of labeled data $|\mathcal{D}|$ when the PDE loss is used on the performance of RPLPO for the NSWE setting. $\mathcal{L}_{D}$, $\mathcal{L}_{P}^{\theta}$ and $\mathcal{L}_{P}^{\phi}$ are applied to train the neural network collaboratively, where $\mathcal{L}_{D}$ is calculated by $u$. Then, we explore if the high-resolution data (when available) can help the training and improve the performance of the proposed RPLPO, although the setting of the proposed RPLPO is under the partial observation.  

For  of $|\mathcal{D}|$, we consider $|\mathcal{D}|\in\{50,100,150,200,250,300,350\}$. The values of the evaluations for RPLPO on NSWE trained are presented in Table \ref{app data labeled data }. As we aim to generalize the model on a variety of trajectories, data number $|\mathcal{D}|$ denotes the number of trajectories beginning from different ICs. The $\mathcal{L}_{D}$ decreases along with the increasing of $|\mathcal{D}|$. In the range of $|\mathcal{D}|$, RPLPO can always bring improvements based on the detailed results in Table \ref{app data labeled data }.
\begin{table*}[h]
\centering
\caption{Relative loss $\mathcal{L}_{D}$ ($\downarrow$) of the data numbers in detail.}
\vskip 0.02in
\label{app data labeled data }
\renewcommand{\arraystretch}{1.2}
\setlength{\tabcolsep}{4pt}
\begin{small}
\begin{sc}
\begin{tabular}{c|ccccccc}
\toprule
Method & 50 & 100 & 150 & 200 & 250 & 300 & 350 \\
\hline
FNO* & 1.28E-1 & 1.07E-1 & 9.14E-2 & 8.03E-2 & 7.34E-2 & 6.41E-2 & 5.59E-2 \\
RPLPO & \textbf{8.81E-2} & \textbf{6.88E-2} & \textbf{5.39E-2} & \textbf{4.48E-2} & 4.01E-2 & \textbf{3.50E-2} & \textbf{3.21E-2} \\
\bottomrule
\end{tabular}
\end{sc}
\end{small}
\end{table*}

We consider three cases when we assume the high-resolution data is available, including 1/3, 1/2, and all partial observation have the corresponding high-resolution data. Then, we apply the high-resolution data as the labels of the encoding module so that it can learn to reconstruct the more accurate and reliable high-resolution state. As the encoding module with high-resolution labels does not need to fine-tune the encoding module, we do the experiment using the RPLPO without fine-tuning by using the high-resolution labels. From \textbf{Table 2 in the main text}, we can see that both the baseline FNO* (without PDE loss) and our proposed RPLPO benefit from the increasing of high-resolution data. It is worth mentioning that the improvements are over 38\% and 12\% when all the high-resolution data are assumed to be available for FNO* and RPLPO, respectively. Moreover, these results also highlight an important aspect of our framework: the ability to use the known PDE as a substitute for high-resolution data in training the encoding module, especially in scenarios where such data is unavailable. The smaller improvement observed in RPLPO compared to the FNO*, when high-resolution data is available, suggests that RPLPO can effectively compensate for the absence of high-resolution data to learn the state.

\subsubsection{Sparsity Level and Irregularity of Observation}
In this part, we first evaluate the different sparsity levels of regular partial observation based on NSWE. Secondly, we use the irregular partial observation data with the same number of values as above to show the effectiveness on irregular data.
\begin{figure}[h]
\centering
\vskip 0.1in
\centerline{\includegraphics[width=0.6\columnwidth]{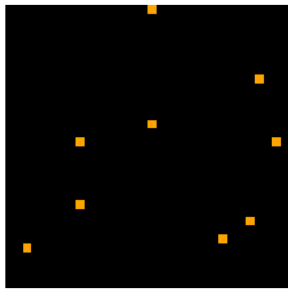}}
\caption{An example of irregular partially observed positions for nine number of points. The orange points are the irregular observations and the black region is unobservable.}
\label{app: irregular mesh}  
\end{figure}
Low sparsity levels have more observed pixels, while high sparsity levels lead to fewer, which means that high sparsity levels have less spatial information. We consider this feature when evaluating our proposed framework. We apply three datasets with observation meshes $x_u\times y_u\in\{3\times3, 5\times5, 7\times7\}$. The above meshes can observe 9, 25, and 49 pixels, respectively, and the sparsity level gradually decreases. The values of the evaluations for RPLPO trained with each size of the $x_u\times y_u$ values are presented in \textbf{Table 4 of the main text}. $\mathcal{L}_{D}$ calculated between prediction and ground truth are worse in high sparsity level like $3\times3$ as the encoding module faces the challenge of reconstructing the high-resolution state only using less spatial knowledge and PDE loss. It has better results when sparsity level is higher with relatively much spatial knowledge. Moreover, In the range of $x_u\times y_u$, RPLPO can always bring improvements based on results. 

The irregular partial observation data are sampled randomly with the number $s=9$. We illustrate an example of the irregular partially observed position in Fig. \ref{app: irregular mesh}. The results of evaluation for RPLPO are shown in \textbf{Table 5 of the main text}. Among all the results, RPLPO has significant improvement over FNO*. The results illustrate that our framework always have similar improvement compared to baselines, whether the observation is regular or irregular.

\subsubsection{Inaccurate PDE and Noisy Data}
After considering the different sparsity levels of partial observation and irregular observation, there are two common challenges encountered in real-world practices: inaccurate PDE and noisy data. Therefore, we consider the above two challenges under partial observation and evaluate the performance of RPLPO. We add the unknown terms as the Gaussian random fields (GRFs) in PDE to represent the inaccuracy and use the data with different levels (10\%, 20\%, and 30\%) Gaussian noise following the previous work \cite{rao2023encoding}. We use the GRFs with mean 0 and std 1, 5, and 10. 

From the results in \textbf{Table 7 of the main text}, RPLPO with accurate PDE and data without noise has the best result. Alone the increasing of data noise percentages and the increasing of scales of GRFs. Among all settings, we can see that RPLPO improves performance against the baseline. It illustrates that our method is also effective in such real-world practices where the PDE is not accurate and the data has noise under partial observation. 

In addition to unknown terms in PDE, there is also a situation in real-world practices where the \textbf{parameters in PDE are inaccurate}. For the NSWE we applied, the most important parameter is the viscosity coefficient $nu$, as shown in Eqn. \ref{eqn.nswe}. The accurate $nu$ is 0.02, and we evaluate the effectiveness of RPLPO on other inaccurate $nu$ in this part. We consider $\nu\in{\{0.01,0.03,0.03,0.05\}}$ and show the results in Table \ref{inaccurate nu}. From the table, we can see that RPLPO improves the model's performance among all parameters, even when $\nu=0.04$ is 2 times the accurate value. Although when $\nu=0.05$ (2.5 times) $\mathcal{L}_{D}$ increases, the performance is still better than baselines (the best in baselines is 6.41E-2).

\begin{table}[t]
    \centering
    \caption{Relative loss $\mathcal{L}_{D}$ ($\downarrow$) of the inaccurate parameter $\nu$.}
    \label{inaccurate nu}
    \renewcommand{\arraystretch}{1.2}
    \setlength{\tabcolsep}{4pt}
    \begin{small}
    \begin{sc}
    \begin{tabular}{c|ccccc}
    \toprule
    $\nu$ & 0.01 & 0.02 & 0.03 & 0.04 & 0.05 \\
    \hline
     $\mathcal{L}_{D}$ & 3.92E-2 & 3.50E-2 & 3.98E-2 & 3.72E-2 & 5.38E-2\\
    \bottomrule
    \end{tabular}
    \end{sc}
    \end{small}
\end{table}

\subsubsection{Zero-shot Super-resolution}

\begin{figure}[h]
\centering
\centerline{\includegraphics[width=0.9\columnwidth]{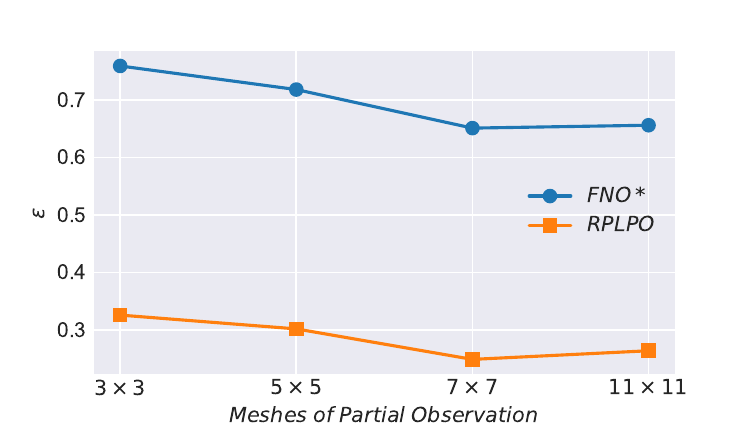}}
\caption{Relative reconstruction error $\epsilon$ ($\downarrow$) about the new resolutions in inference (zero-shot super-resolution performance).}
\label{zero-shot sr}  
\end{figure}

The original U-Net encoding module requires consistent partial observation meshes between inference and training. The goal of our work is to develop a training framework that is not limited to the specific encoding module. In this experiment, we replace the U-Net with a Transformer as an encoding module and release the zero-shot super-resolution capacity following the method in MAgNet \citep{boussif2022magnet}. MAgNet is a mesh-based neural operator that enables zero-shot generalization to new meshes. We apply the nearest neighbors interpolation as same as MAgNet in the encoding module. We train the model on $7\times7$ partial observation, and test zero-shot super-resolution on ${3\times3, 5\times5, 7\times7, 11\times11}$. We use the reconstruction error $\epsilon$ to measure the zero-shot super-resolution performance. The results presented in the Fig. \ref{zero-shot sr} shows that the RPLPO outperforms the FNO* method, whether the inference size of the partial observation is. It indicates that the using of PDE loss can significantly improve the zero-shot super-resolution of the encoding module, by ensuring that the output adheres to the underlying PDE, especially when high-resolution data is not available.

\subsubsection{Physics Metrics} Even if the data loss is a widely used metric in quantifying the performance, it only illustrates the dfferences on pixel. We additionally evaluate our framework on three physics metrics: divergence, Turbulence Kinetic Energy (TKE), and energy spectrum of NSE following \citet{wang2020towards}. 

\begin{table}[t]
    \centering
    \caption{Relative error ($\downarrow$) of divergence, TKE and energy spectrum.}
    \label{physics metrics available result}
    \renewcommand{\arraystretch}{1.2}
    \setlength{\tabcolsep}{4pt}
    \begin{small}
    \begin{sc}
    \begin{tabular}{c|cc}
    \toprule
    Physics metrics & FNO* & \textbf{RPLPO} \\
    \hline
    Divergence error & 2.24E-2 & \textbf{1.68E-2}\\
    TKE error & 8.17E-4 & \textbf{3.89E-5} \\
    Energy spectrum error & 1.63E-3 & \textbf{7.78E-5} \\
    \bottomrule
    \end{tabular}
    \end{sc}
    \end{small}
\end{table}

Divergence: Since we investigate incompressible turbulent flows in this work, which means the divergence, $\nabla \cdotp (u_x, u_y)$, at each pixel should be 0, we use the average of absolute divergence over all pixels at each prediction step as an additional evaluation metric

TKE: In fluid dynamics, turbulence kinetic energy is the mean kinetic energy per unit mass associated with eddies in turbulent flow. Physically, the turbulence kinetic energy is characterised by measured root mean square velocity fluctuations.
\begin{equation}\small
\begin{aligned} 
(\overline{(u_{x}^{'})^2} + \overline{(u_{y}^{'})^2})/2, \overline{(u_{x}^{'})^2}=\frac{1}{T}\displaystyle\sum_{t=0}^{T}(u_x(t)-\overline{u_x})^2,
\label{eqn.tke}
\end{aligned}
\normalsize
\end{equation}
where $t$ is the time step. 

Energy Spectrum The energy spectrum of turbulence, $E(k)$, is related to the mean turbulence kinetic energy as
\begin{equation}\small
\begin{aligned} 
\int_{0}^{\infty }E(k)dk=(\overline{(u_{x}^{'})^2} + \overline{(u_{y}^{'})^2}),
\label{eqn.tke}
\end{aligned}
\normalsize
\end{equation}
where $k$ is the wavenumber, the spatial frequency in 2D Fourier domain. We calculate the Energy Spectrum on the Fourier transformation of the Turbulence Kinetic Energy fields.

We show their relative error between the predicted value and the ground truth in the Table \ref{physics metrics available result}. Among the physics metrics, our proposed RPLPO show the significnat improvement against the baseline, which means that RPLPO can learn better physical meaning and preserve desired physical quanti ties.


\subsubsection{Network Architecture of Encoding Module} The aim of this work is to propose a learning framework that can apply PDE loss on partial observation to improve generalization capacity. In practical implementation, the encoding process is similar to the image processing task, which aims to extract features from partial observation input. In response to this, we follow the works and employ the U-Net architecture \citep{ronneberger2015u} as the encoding module, a proven model well-suited for tasks akin to super-resolution \citep{esmaeilzadeh2020meshfreeflownet}. We replace the U-Net with the Transformer to illustrate the robustness of the network architecture of encoding module, based on the NSWE setting. We employ the official implementation of the Transformer in ViT \citep{dosovitskiy2020image}, which is a deep learning architecture for computer vision tasks that leverages the Transformer model's self-attention mechanism to process images as sequences of tokens. The results are shown in \textbf{Table 8 of the main text}. We can see that RPLPO can still have the improvement compared with the data-driven manner FNO* by using the PDE loss, especially when it is fine-tuned using the unlabeled partial observation after replacing the encoding module from U-Net to Transformer. 

To verify the necessity and effectiveness of the proposed encoding module and physics-informed learning manner, we also design experiments to replace the PDE loss trained encoding module to some interpolation methods, including nearest neighbors interpolation, bilinear interpolation and bicubic interpolation. As shown in \textbf{Table 9 of the main text}, we can see that our proposed encoding module trained by PDE loss can learn the more accurate high-resolution state than all interpolation methods significantly.

\subsubsection{Ablation Study on Hyperparameters} 
We evaluate the sensitive of the weights given $\mathcal{L}_{P}^{\theta}$ and $\mathcal{L}_{P}^{\phi}$, the input covering lengths of most recent consecutive observations, and the coefficients in the two-stage fine-tuning period on the performance of RPLPO for NSWE setting. As the results of FNO* are relatively better than other baselines, serving as a representative for them, we focus on the improvement between FNO* and RPLPO in the following ablation studies. 

For weights of PDE loss, there are $\mathcal{L}_{D}$, $\mathcal{L}_{P}^{\theta}$ and $\mathcal{L}_{P}^{\phi}$, where $\mathcal{L}_{P}^{\phi}$ and $\mathcal{L}_{P}^{\theta}$ are weighted with coefficient $\gamma$. Accordingly, we study the influence of the hyperparameter $\gamma$ in the loss function $\mathcal{L}_{PI}$. For this purpose, we consider $\gamma\in$ \{0, 1E-3, 5E-2, 1E-1, 2E-1, 1\}. The values of the evaluations for RPLPO based on NSWE trained with each of the $\gamma$ values are presented in Table \ref{app weight result}. $\gamma=0$ indicates the FNO* depends on the $\mathcal{L}_{D}$, and $\mathcal{L}_{P}^{\phi}$ and $\mathcal{L}_{P}^{\theta}$ are not accounted for. As presented in Table \ref{app weight result}, the best performance on the test set is achieved for a loss function with weighting coefficients of $\gamma$ = 1E-1, 2E-1. Allover this work, we refer to the weight coefficient with $\gamma$ = 1E-1. A model that is trained by data-driven, which only focuses on the data (i.e., uses $\mathcal{L}_{D}$ only) and does not account for the PDE loss (i.e., ignores $\mathcal{L}_{P}^{\phi}$ and $\mathcal{L}_{P}^{\theta}$) underperform than that of RPLPO. On the other hand, models trained with a significant focus on the PDE loss only (i.e., large $\gamma$) also underperform. In general, a balance between the focus of the RPLPO on the data and the PDE losses leads to an optimal performance. 
\begin{table*}[h]
\centering
\caption{Relative loss $\mathcal{L}_{D}$ ($\downarrow$) of different weight of PDE loss in detail.}
\vskip 0.02in
\label{app weight result}
\renewcommand{\arraystretch}{1.2}
\setlength{\tabcolsep}{4pt}
\begin{small}
\begin{sc}
\begin{tabular}{c|cccccc}
\toprule
Method & 0 & 1E-3 & 5E-2 & 1E-1 & 2E-1 & 1 \\
\hline
\textbf{RPLPO} & 6.41E-2 & 4.16E-2 & 3.61E-2 & \textbf{3.50E-2} & 3.52E-2 & 3.49E-2 \\
\bottomrule
\end{tabular}
\end{sc}
\end{small}
\end{table*}

When two PDE lossess ($\mathcal{L}_{P}^{\theta}$, $\mathcal{L}_{P}^{\phi}$) have different $\gamma$. The results are shown in Table \ref{different weight table}, which illustrate our method is robust to different $\gamma$.

\begin{table}[H]
    \centering
    \caption{Relative error ($\downarrow$) of different $\gamma$.}
    \label{different weight table}
    \renewcommand{\arraystretch}{1.2}
    \setlength{\tabcolsep}{4pt}
    \begin{small}
    \begin{sc}
    \begin{tabular}{c|c}
    \toprule
    $\gamma$ & $\mathcal{L}_{D}$ \\
    \hline
    (1E-1,1E-1) & 3.50E-2 \\
    (1E-1,2E-1) & 3.58E-4 \\
    (2E-1,1E-1) & 3.38E-3 \\
    \bottomrule
    \end{tabular}
    \end{sc}
    \end{small}
\end{table}

For lengths of recent consecutive observations as input, there are recent consecutive observations input to the model, balanced with length coefficient $n$, fed to the encoding module to make up for the lack of spatial information with the of temporal information, drawing inspiration from PDEs formulation and FDM, where there exists an equivalence relation between temporal and spatial differences, allowing for interconversion between them. Accordingly, we study the influence of the hyperparameter $n$, deciding the temporal features carried by input data on the performance of RPLPO. For this purpose, we consider $n\in\{1,4,6,8\}$. The values of the evaluations for RPLPO based on NSWE trained with each of the $n$ values are presented in Table \ref{app history result}. $n=1$ indicates the input only has the current partial observation, and the most recent consecutive observations are not accounted for. As presented in Table \ref{app history result}, when covering the coefficient of $n=4$, the evaluation on the test set has achieved a good result. A model that is trained by input without recent consecutive observations faces a challenge when the encoding module reconstruct high-resolution state leveraging insufficient spatial information by PDE loss. Allover this work, we refer to this length coefficient $n=4$. 
\begin{table}[h]
\centering
\caption{Relative loss $\mathcal{L}_{D}$ ($\downarrow$) of length of recent consecutive observations as input in detail.}
\vskip 0.02in
\label{app history result}
\renewcommand{\arraystretch}{1.2}
\setlength{\tabcolsep}{4pt}
\begin{small}
\begin{sc}
\begin{tabular}{c|cccc}
\toprule
Method & 1 & 4 & 6 & 8 \\
\hline
FNO* & 1.16E-1 & 6.41E-2 & 8.43E-2 & 1.10E-1 \\
\textbf{RPLPO} & \textbf{8.94E-2} & \textbf{3.50E-2} & \textbf{4.00E-2} & \textbf{4.94E-2} \\
\bottomrule
\end{tabular}
\end{sc}
\end{small}
\end{table}


For the coefficients in the two-stage fine-tuning period, in the first tuning stage, unlabeled data are applied to tune the transition module using only $\mathcal{L}_{P}^{\phi}$ without $\mathcal{L}_{D}$. Then, we apply $\mathcal{L}_{D}$ to tune the encoding module independently in the second stage, where $\mathcal{L}_{D}$ is calculated based on the original training set $\mathcal{D}$ without introducing new labeled data. The steps of the two stages are controlled by the coefficients $m_1$ and $m_2$ to make RPLPO achieve better performance in evaluation. Accordingly, we study the impact of the hyperparameters $m_1$ and $m_2$. For this purpose, we consider $m_1\in\{5,10\}$ and $m_2\in\{5,10\}$. The values of the evaluations for RPLPO based on NSWE trained with each of the $m_1$ and $m_2$ values are presented in Table \ref{app finetune result}. As presented in Table \ref{app finetune result}, when the coefficients $m_1=10$ and $m_2=10$, the evaluation on the test set has achieved a good result. A model with less $m_1$ and $m_2$ fails to modify the performance significantly, while a model with more steps causes too much deterioration in the first stage. Allover this work, we refer to this optimum coefficients $m_1=10$ and $m_2=10$. Another coefficient is the gap between each two-stage fine-tuning period. We further study the impact of gap coefficient $q$. For this purpose, we consider $q\in\{50,100,200,300\}$ and control the same training epoch of the base-training period. The values of the evaluations for RPLPO on NSWE trained with each of the $q$ values are presented in Table \ref{app finetune result}. When the gap coefficient $q=100$, the evaluation of the test set has achieved the best result. A model with less $q$ tunes too frequently, causing instability of base-training, while a model with more $q$ does not have an obvious impact on results. Allover this work, we refer to this optimum gap coefficient $q=100$.
\begin{table*}[h]
\centering
\caption{Relative loss $\mathcal{L}_{D}$ ($\downarrow$) of coefficient in the two-stage fine-tuning period in detail.}
\vskip 0.02in
\label{app finetune result}
\renewcommand{\arraystretch}{1.2}
\setlength{\tabcolsep}{4pt}
\begin{small}
\begin{sc}
\begin{tabular}{c|cccccc}
\toprule
Method & 50/10/10 & 100/5/5 & 100/10/5 & 100/10/10 & 200/10/10 & 300/10/10 \\
\hline
\textbf{FNO*} & 6.41E-2 & 6.41E-2 & 6.41E-2 & 6.41E-2 & 6.41E-2 & 6.41E-2 \\
\textbf{RPLPO} & \textbf{3.58E-2} & \textbf{3.55E-2} & \textbf{3.50E-2} & \textbf{3.53E-2} & \textbf{3.58E-2} & \textbf{3.64E-2} \\
\bottomrule
\end{tabular}
\end{sc}
\end{small}
\end{table*}


\subsubsection{Necessity of Fine-tuning} 
The two-stage fine-tuning period is an important part of RPLPO as data scarcity tends to cause generalization issues, especially when data acquisition is costly, and only partial observation is available. As shown in Fig. 1, in this period, transition PDE loss $\mathcal{L}_{P}^{\phi}$ (calculated using unlabeled data) is used to train the transition module and we stop gradient of the encoding module, then the data loss $\mathcal{L}_{D}$ (calculated using the original labeled data) is used to train the encoding module and we stop gradient of the transition module. We conduct the ablation study on the relationship between the data and the effect of fine-tuning. In \textbf{Table 2 of the main text}, the RPLPO w/o FT means RPLPO is not trained by the two-stage fine-tuning period. RPLPO has greater improvement than RPLPO w/o FT when data is limited, as leveraging the unlabeled data provides additional information.

\subsubsection{Impact of PDE Loss on Computational Cost and GPU Usage} 
As the computation of PDE loss leads to more computational cost and GPU memory usage than the data-driven manner, we study the increasing of computational cost and GPU memory usage when the output resolution of the encoding module $E_\theta$ increases based on the NSWE setting.

We employ the hyperparameters in the same way as the experiment in Technical Appendix Model Architecture, shown in Table \ref{app hyperparameters used for structure} and Table \ref{app hyperparameters used for training}, and calculate the computational time and GPU memory usage of the model's inference and batch training. The models are trained and evaluated on a single Nvidia V100 16GB GPU. We compare them of the proposed RPLPO with baseline FNO* to illustrate the variance between physics-informed training manner and data-driven manner. From \textbf{Table 6 of the main text}, we can see no matter how much the resolution is, the inference of FNO* and RPLPO always have similar cost and GPU memory as they have the same model structure. During the training, these of RPLPO are higher than that of FNO* as they have different training loss functions, and RPLPO is required to calculate the PDE loss and derivatives of them with respect to two modules, but they are still in the reasonable range.

\subsubsection{Cost on Transition Module} 
We evaluate the different computational cost of our transition module against the numerical solver to verify the necessity of our proposed component of RPLPO. We calculate the computational time of a step forward by our transition module and a step computation of the numerical solver by FDM, to illustrate the computational cost. We do the experiments on different resolutions based on NSWE, including ${32\times32, 48\times48, 64\times64}$. The results are shown in \textbf{Table 10 of the main text}. We can see that the time taken for a single computation using a numerical solver is more than 5 times that of a step forward with a neural network. Furthermore, as the resolution increases, the computational cost of using a neural network does not significantly change, as only the input and output layers are affected, with small changes in the hidden layers. These results indicate that, in terms of computational cost, whether in training or inference, employing a neural network in the transition module offers considerable advantages.

\subsubsection{Ablation Study on the Types of Data Loss}\label{ref:detail of data loss}
\begin{table}[ht]
\centering
\caption{Relative loss $\mathcal{L}_{D}$ ($\downarrow$) of RPLPO with different types of loss.}
\vskip 0.02in
\label{app loss}
\renewcommand{\arraystretch}{1.2}
\setlength{\tabcolsep}{2.6pt}
\begin{small}
\begin{sc}
\begin{tabular}{c|ccc}
\toprule
Method & Relative $l_2$ & $l_2$ & $l_1$ \\
\hline
FNO* & 6.41E-2 & 7.90E-2 & 7.53E-2 \\
\textbf{RPLPO} & 3.59E-2 & \textbf{6.18E-2} & 6.80E-2 \\
\bottomrule
\end{tabular}
\end{sc}
\end{small}
\end{table}
We conducted the ablation study using $l_1$, $l_2$ and relative $l_2$ loss during training on NSWE setting. As shown in Table \ref{app loss}, the models trained with the relative $l_2$ loss perform better than $l_1$ and $l_2$ loss. \citet{kovachki2023neural} has discovered training with the relative loss results in around half the testing error rate compared to training with the $l_2$ loss. Moreover, no matter what the loss function is, our framework can always improve the performance by using PDE loss.

\subsubsection{Ablation Study on the Time-stepping Methods and Finite-difference Schemes}\label{ref:RK4 and FDM}
\begin{table}[ht]
\centering
\caption{Relative loss $\mathcal{L}_{D}$ ($\downarrow$) of RPLPO with different time-stepping methods in PDE loss.}
\vskip 0.02in
\label{app rk4 ablation}
\renewcommand{\arraystretch}{1.2}
\setlength{\tabcolsep}{2.6pt}
\begin{small}
\begin{sc}
\begin{tabular}{c|cccc}
\toprule
Methods & RK2 & RK3 & RK4 & RK5 \\
\hline
FNO* & 6.41E-2 & 6.41E-2 & 6.41E-2 & 6.41E-2 \\
\textbf{RPLPO} & 3.59E-2 & 3.52E-2 & \textbf{3.50E-2} & 3.51E-2\\
\bottomrule
\end{tabular}
\end{sc}
\end{small}
\end{table}

We conducted experiments on the NSWE setting that changed the time-stepping methods to the second-order, third-order, and fifth-order Runge-Kutta methods when calculating the PDE loss. We can see from the Table \ref{app rk4 ablation} that regardless of the order of Runge-Kutta method, our proposed framework have significant improvements. Among all types of time-stepping methods, the RK4 has a relative small $\mathcal{L}_{D}$ and we selected it in our main text.

\begin{table}[ht]
\centering
\caption{Relative loss $\mathcal{L}_{D}$ ($\downarrow$) of RPLPO with different difference schemes in PDE loss.}
\vskip 0.02in
\label{app FDM ablation}
\renewcommand{\arraystretch}{1.2}
\setlength{\tabcolsep}{2.6pt}
\begin{small}
\begin{sc}
\begin{tabular}{c|ccc}
\toprule
Methods & second & fourth & sixth \\
\hline
FNO* & 6.41E-2 & 6.41E-2 & 6.41E-2 \\
\textbf{RPLPO} & 3.51E-2 & \textbf{3.50E-2} & 3.52E-2\\
\bottomrule
\end{tabular}
\end{sc}
\end{small}
\end{table}

We conducted an ablation study that change the fourth-order central difference scheme to the second-order and sixth-order central difference scheme when calculating the PDE loss. The performance of relative data loss is shown in the Table \ref{app FDM ablation}. We can see that regardless of the order of FDM, our proposed framework can bring significant improvements. The result also proves that our framework is robust to calculating errors in PDE loss and can apply FDMs of different orders. Among all schemes, the fourth-order has a relative small $\mathcal{L}_{D}$ and we selected it in our main text.

\subsubsection{Ablation Study on the Upscale Factor of the Encoding Module}
We conducted the ablation study to show the different upscale factor of the encoding module to other high resolutions and also show the reliability of PDE loss on the selected high-resolution. We apply a higher resolution $128\times 128$ and compare its result with our selected resolution $32\times 32$ in the main experiments on two datasets of the shallow water equation (LSWE and NSWE). The results are shown in Table \ref{app high-reso}.

\begin{table}[H]
\centering
\caption{Relative loss $\mathcal{L}_{D}$ ($\downarrow$) of different upscale factors.}
\vskip 0.02in
\label{app high-reso}
\renewcommand{\arraystretch}{1.4}
\setlength{\tabcolsep}{6pt}
\begin{small}
\begin{sc}
\begin{tabular}{c|cc}
\toprule
Settings & $32\times 32$ & $128\times 128$ \\
\hline
LSWE & \textbf{2.44E-2} & 2.49E-2  \\
NSWE & 3.50E-2 & \textbf{3.21E-2} \\
\bottomrule
\end{tabular}
\end{sc}
\end{small}
\end{table}

From the table, we can draw the conclusion that the high resolution we selected is reliable, and our framework can be applied to higher resolutions. In the table, we can see that the relative loss of $128\times 128$ on LSWE is slightly higher than that of the result of $32\times 32$ and that on NSWE is slightly lower than the result of $32\times 32$. We consider that there is a trade-off: increasing the resolution will make the PDE loss more accurate, but it will pose a challenge to the encoding module as higher dimension state is required to be reconstructed.

\section{Related Works}  \label{app: related work}
\textbf{Physics-informed Machine Learning:} There are two categories of physics-informed machine learning. The first one is the data-driven method using the dataset collected from solvers or experiments, like the neural operators \citep{lu2019deeponet, li2020neural, li2020fourier, gupta2021multiwavelet, boussif2022magnet, yin2023continuous, iakovlev2023learning, hansen2023learning, chen2022crom}. Another one is Physics-Informed Neural Networks (PINNs) \citep{raissi2019physics, yang2021b, cai2021physics, karniadakis2021physics} for training physics-based loss to solve equations. Both approaches have disadvantages. On the one hand, neural operators require data, however, data generation might require the enormous cost of the expensive solver and experiment. On the other hand, PINNs do not mandate the input of data, which tends to exhibit limitations, particularly in the context of multi-scale dynamic systems, attributable to the complexities of optimization \citep{li2020neural}. In order to overcome the above challenges, physics-informed operator learning has been proposed in PI-DeepONet \citep{wang2021learning, goswami2022physics_1} and PINO \citep{li2021physics} that reduce the need for data by using PDE loss and learn an operator to generalize multi-scale dynamics.
These works leverage PDE loss in constructing PDE loss or network structure, thereby modeling or solving PDE dynamics. The key to their success is the incorporation of accurate PDE, including high-resolution data or formulas of PDEs. However, they cannot be applied to learn partial observation directly, as their physics-informed training manner brings significant bias when calculating by partial observation. 

\textbf{High-resolution State Reconstruction:} This task aims to reconstruct a high-resolution state from its partial observation, also known as super-resolution in some works \cite{zhu2020beyond, li2021model}. These tasks are focused on two domains: computer vision (CV) and physical systems. In CV, the pioneering study in \citet{dong2014learning} was among the first to utilize deep learning for this task. Subsequent to this, numerous deep learning-based models \citep{lim2017enhanced, soh2019natural, nazeri2019edge, zhao2020efficient}, and generative models \citep{ledig2017photo, liu2021variational, gao2023implicit} emerged to enhance the performance on such tasks. For physical systems, the task has attracted more and more attention \citep{ren2023physr, ren2023superbench}, which aims to use PDE loss in the model \citep{wang2020physics, esmaeilzadeh2020meshfreeflownet, fathi2020super, ren2023physr, jangid2022adaptable, shu2023physics} or develop the physics-informed super-resolution models without data  \citep{gao2021super,kelshaw2022physics,zayats2022super}. Both methods face challenges in terms of data requirements and generalization despite their individual merits. On the one hand, the works in CV and most of the work in physical systems require high-resolution state used in supervised learning. On the other hand, other works in physical systems only rely on PDE loss, resulting relative larger reconstruction error, and hard to model the unsteady dynamics\citep{gao2021super} without leveraging data information. Our proposed RPLPO applies physics-informed training to learn the high-resolution state via partial observation without high-resolution labels and firstly embeds it to a prediction task further instead of using reconstruction as an end-to-end task like the above works. Although the reconstruction using only PDE loss is still not accurate enough, these high-resolution reconstructions that more in line with PDE can reduce bias when handling partial observation to overcome the challenge in physics-informed neural operator. When we face problems of data scarcity and partially observable nature of observation, it can improve the generalization of predictions in physical systems modeling. There is a work\citep{rao2023encoding} that looks most related to our work; in this work, Chengpeng Rao, \textit{et al.} also considered the high-resolution state reconstruction and the time-series prediction problem, but unlike our work, it is similar to the neural PDE solvers like PINNs,
rather than operator learning. It cannot generalize different initial conditions and trajectories. 
We have also leveraged this work as one of our baselines.

\section{Multi-step Prediction}  \label{app multi step result}

The additional results of the multi-step predictions for five benchmarks introduced in the main text is demonstrated in Tables \ref{tab.4}.

\section{Result Visualization} \label{app visual}
\begin{figure*}[tb]
\centering
\centerline{\includegraphics[width=\textwidth]{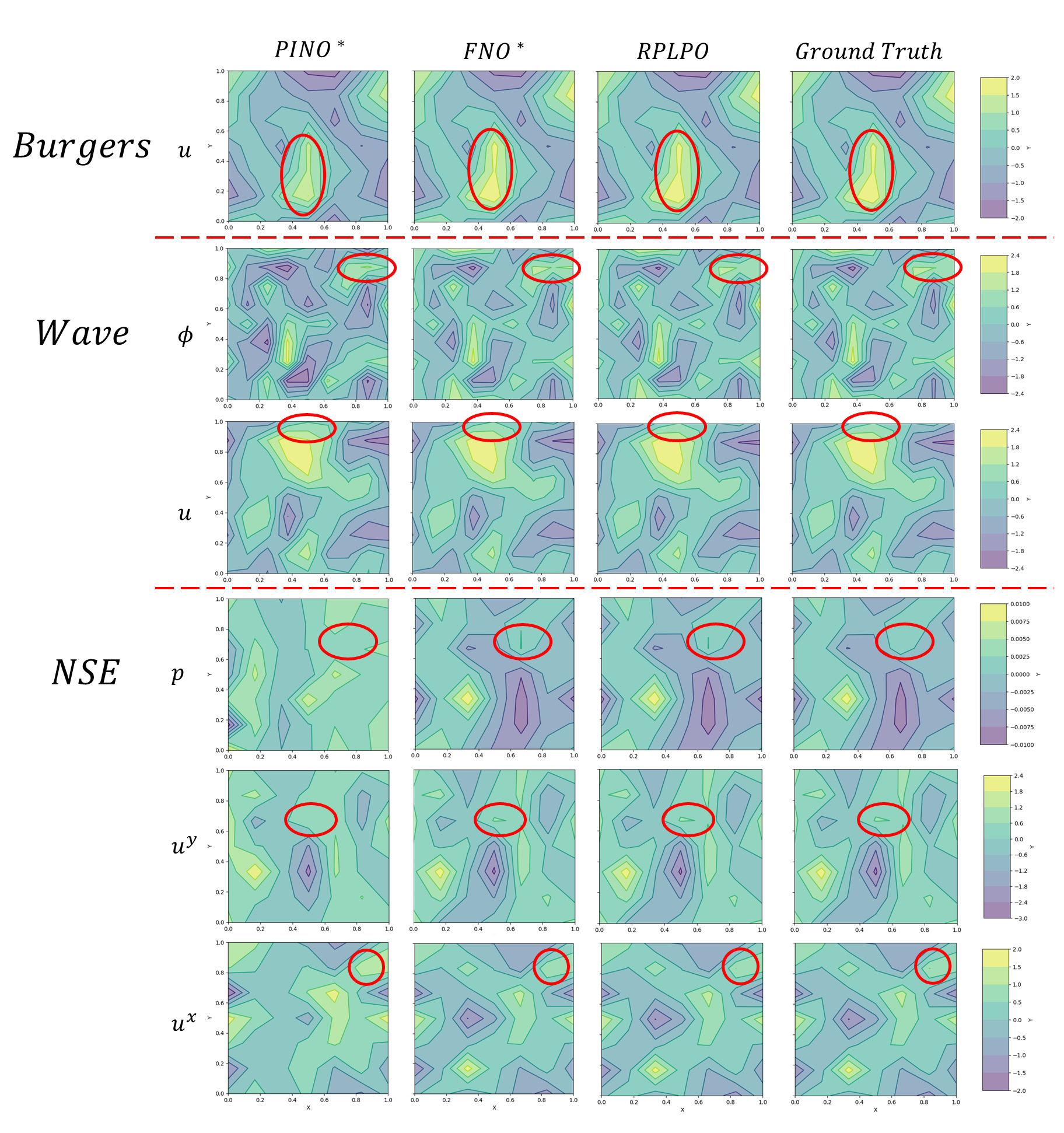}}
\caption{\textbf{Visualization of predictions in Burgers (\textbf{Top}), Wave (\textbf{Middle}) and NSE (\textbf{Bottom}) experiments, using PINO*, FNO*, RPLPO, and ground truth.} We use red circles to illustrate one of the improvements using RPLPO. \textbf{Burgers} $u$ denotes the field of velocity. \textbf{Wave} $\phi$ denotes the field of velocity potential and $u$ denotes field of velocity. \textbf{NSE} $p$ denotes the field of pressure, $u^x$ and $u^y$ denote the fields of velocity in $x$ and $y$ directions. We can see that our RPLPO (third column) learns details better than PINO* and FNO*.}
\label{app_plot_1}
\end{figure*}

\begin{figure*}[tb]
\centerline{\includegraphics[width=\textwidth]{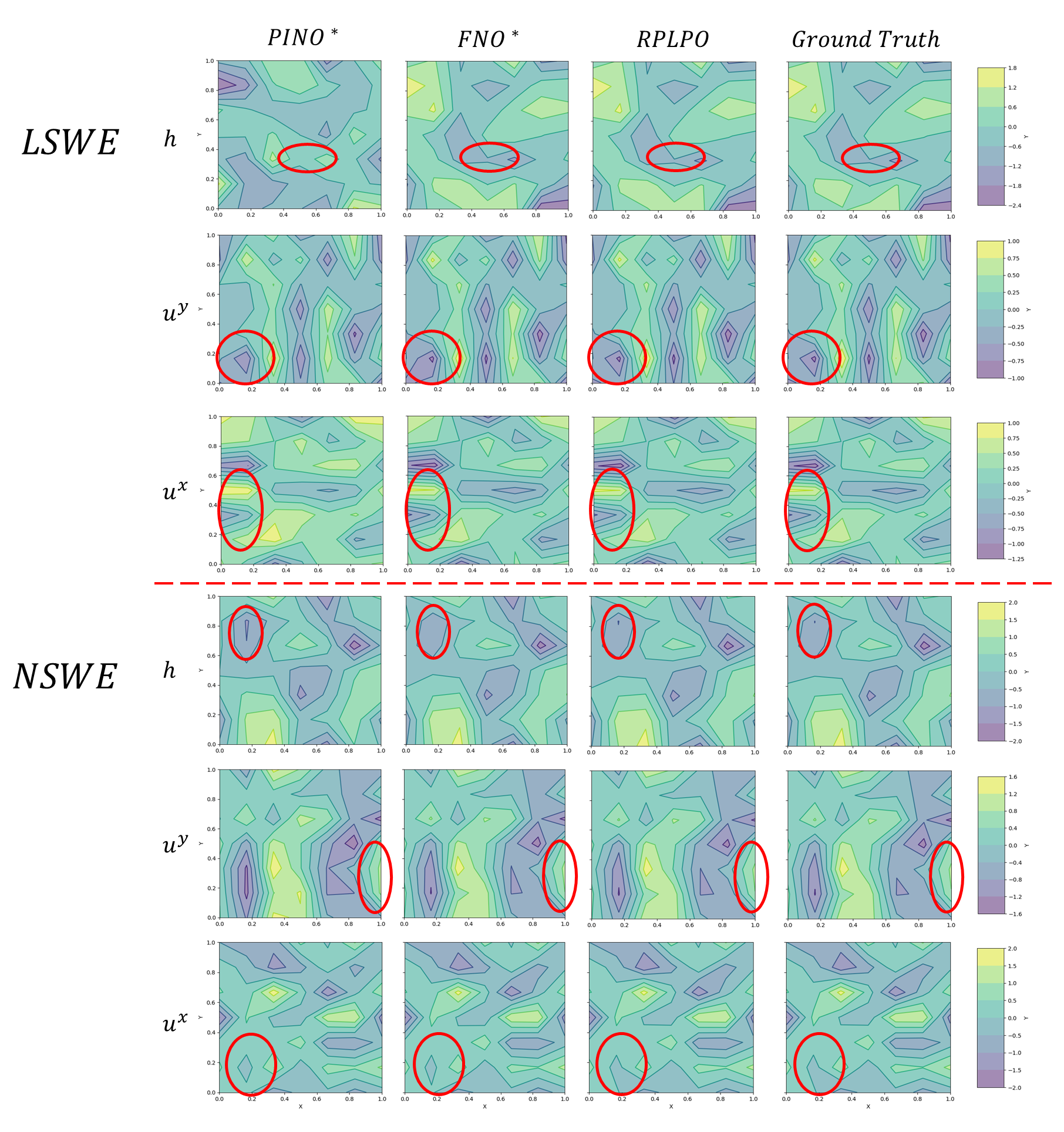}}
\caption{\textbf{Visualization of predictions in LSWE (\textbf{Top}) and NSWE (\textbf{Bottom}) experiments, using PINO*, FNO*, RPLPO, and ground truth.} We use red circles to illustrate one of the improvements using RPLPO. $h$ denotes the field of fluid column height, $u^x$ and $u^y$ denote the fields of velocity in $x$ and $y$ directions. We can see that our RPLPO (third column) learns details better than PINO* and FNO*.
}
\label{app_plot_2}
\end{figure*}

In this section, we provide the result visualizations for the Burgers, Wave, NSE, LSWE, NSWE. The results of PINO*, FNO*, RPLPO, and ground truth are shown in Fig. \ref{app_plot_1} and Fig. \ref{app_plot_2}. RPLPO (third column) can learn more accurate details than PINO* and FNO*. We use red circles to illustrate one of the improvements using RPLPO. For the Burgers, there is an obvious improvement on the width of the middle of shape in the circle. For the Wave, there is an obvious improvement on $\phi$, like the different number of triangles in the circle, and the slight improvement on $u$ located near the edge. For the NSE, there are the improvements on the length of line in the circle on $p$, and the size of triangle on $u_y$, and the dot in the center of circle on $u_x$. For the LSWE, there is the slight improvement on $h$, like the width in the middle of the saddle shape, an obvious improvement on $u^y$, like different shapes, and the slight improvement on $u^x$ like the size of the triangle at the edge. For the NSWE, there is an obvious improvement on $h$, like the middle dot, the slight improvement on $u^y$, like the width of the triangle at the edge, and an obvious improvement on $u^x$, like the link at the bottom. Please note that the visualizations are only a randomly selected example, more improvements have been observed in the experiments.

\section{Impact Statements and Subsequent Applications} \label{app impact}
Leveraging PDE loss has been proven to be an effective technique for improving deep model's generalization in physical systems modeling. Our proposed framework is expected to re-enable the application of PDE loss under broader situations in the real world to significantly improve the model's generalization capacity. It has a broad impact on modeling in science and engineering fields.

As we mentioned in the abstract, considering that most real-world problems are observed from sensors, and sensors are typically sparse, whether it's for prediction or control problems \cite{desouky2019wave, jamei2022designing}, the information we can obtain is limited to these sensor observations. Therefore, control algorithms that can be applied in real-world scenarios \cite{paris2021robust, li2022machine, castellanos2022machine} rely on partial observations. Our proposed framework can be applied to predict the future situation of the above problems and use it in subsequent control tasks, facilitating model-based control.

\section{Limitations} \label{app limit}
Our proposed RPLPO has several limitations, presenting opportunities for future work. Firstly, although we simulate real-world challenges with numerical data, such as sparse partial observation, inaccurate PDE, and noisy data, our exploration of data in real-world practice is limited due to the lack of real-world benchmarks. Secondly, we mainly tested U-Net, Transformer as encoding modules, and FNO as the transition module. Future work could explore the efficacy of alternative network architectures. For instances, using GNN or GraphFormers to handle the more complex irregular observation, and using FNO or ViT as the encoding module to achieve the discretization-invariant, but it is not the focus of this paper. We leave it as the future work. 

\begin{sidewaystable*}
\centering
\caption{\textbf{Performance of RPLPO and baselines in the multi-step prediction of Burgers, Wave, NSE, LSWE, and NSWE, measured by relative loss $\mathcal{L}_{D}$ ($\downarrow$).} The first row is the steps predicted from the 1\textsuperscript{st} step to 10\textsuperscript{th} step. We can see that our proposed RPLPO consistently has the best results shown as \textbf{bold}.}
\vskip 0.02in
\label{tab.4}
\renewcommand{\arraystretch}{1.5} 
\begin{small}
\begin{sc}
\begin{tabular}{>{\centering\arraybackslash}p{2.5cm}|>{\centering\arraybackslash}p{2.5cm}|cccccccccc}
\hline
\toprule
Benchmark & Method & 1 & 2 & 3 & 4 & 5 & 6 & 7 & 8 & 9 & 10 \\
\hline
\multirow{5}{*}{Burgers} & PIDL & 1.33E-1 & 2.50E-1 & 3.53E-1 & 4.44E-1 & 5.25E-1 & 5.97E-1 & 6.64E-1 & 7.28E-1 & 7.90E-1 & 8.52E-1 \\
 & PINO* & 5.06E-2 & 9.64E-2 & 1.38E-1 & 1.77E-1 & 2.12E-1 & 2.46E-1 & 2.78E-1 & 3.09E-1 & 3.34E-1 & 3.70E-1 \\
 & FNO & 2.02E-2 & 4.35E-2 & 7.04E-2 & 9.73E-1 & 1.24E-1 & 1.50E-1 & 1.75E-1 & 2.00E-1 & 2.24E-1 & 2.48E-1 \\
 & FNO* & 1.70E-2 & 3.29E-2 & 4.79E-2 & 6.19E-2 & 7.51E-2 & 8.76E-2 & 9.93E-2 & 1.10E-1 & 1.21E-1 & 1.31E-1 \\
 & \textbf{RPLPO} & \textbf{1.43E-2} & \textbf{2.75E-2} & \textbf{3.98E-2} & \textbf{5.12E-2} & \textbf{6.18E-2} & \textbf{7.15E-2} & \textbf{8.06E-2} & \textbf{8.91E-2} & \textbf{9.69E-2} & \textbf{1.04E-1} \\
\hline
\multirow{5}{*}{Wave} & PIDL & 1.36 & 2.41 & 3.27 & 4.10 & 4.76 & 5.36 & 6.07 & 6.68 & 6.96 & 7.09 \\
 & PINO* & 1.18 & 2.00 & 2.54 & 2.91 & 3.01 & 2.94 & 2.84 & 2.80 & 3.01 & 3.46 \\
 & FNO & 1.29E-1 & 1.36E-1 & 1.51E-1 & 1.76E-1 & 1.90E-1 & 2.00E-1 & 2.23E-1 & 2.73E-1 & 3.05E-1 & 3.34E-1 \\
 & FNO* & 4.36E-2 & 7.99E-2 & 1.02E-1 & 1.15E-1 & 1.34E-1 & 1.58E-1 & 1.79E-1 & 2.04E-1 & 2.31E-1 & 2.53E-1 \\
 & \textbf{RPLPO} & \textbf{2.60E-2} & \textbf{4.87E-2} & \textbf{6.38E-2} & \textbf{7.57E-2} & \textbf{9.51E-2} & \textbf{1.19E-1} & \textbf{1.45E-1} & \textbf{1.70E-1} & \textbf{1.91E-1} & \textbf{2.07E-1} \\
\hline
\multirow{5}{*}{NSE} & PIDL & 5.97E-1 & 8.24E-1 & 1.15 & 1.74 & 2.76 & 2.34 & 3.17 & 3.89 & 4.64 & 5.62 \\
 & PINO* & 3.98E-1 & 6.26E-1 & 1.00 & 1.62 & 2.43 & 1.96 & 2.48 & 3.18 & 4.06 & 5.17S \\
 & FNO & 8.88E-2 & 1.37E-1 & 1.70E-1 & 1.95E-1 & 2.16E-1 & 2.35E-1 & 2.53E-1 & 2.70E-1 & 2.87E-1 & 3.03E-1 \\
 & FNO* & 3.23E-2 & 6.30E-2 & 9.29E-1 & 1.22E-1 & 1.49E-1 & 1.76E-1 & 2.02E-1 & 2.27E-1 & 2.51E-1 & 2.75E-1 \\
 & \textbf{RPLPO} & \textbf{2.72E-2} & \textbf{5.31E-2} & \textbf{7.89E-2} & \textbf{1.04E-1} & \textbf{1.28E-1} & \textbf{1.52E-1} & \textbf{1.75E-1} & \textbf{1.99E-1} & \textbf{2.23E-1} & \textbf{2.49E-1} \\
\hline
\multirow{5}{*}{LSWE} & PIDL & 2.84 & 6.10 & 10.10 & 15.17 & 21.58 & 29.33 & 37.94 & 46.31 & 53.28 & 58.40 \\

& PINO* & 2.83 & 6.06 & 9.85 & 14.42 & 19.91 & 26.19 & 32.65 & 38.25 & 42.07 & 43.90 \\

& FNO & 4.13E-1 & 4.24E-1 & 4.32E-1 & 4.43E-1 & 4.58E-1 & 4.76E-1 & 4.99E-1 & 5.25E-1 & 5.55E-1 & 5.85E-1 \\

& FNO* & 2.18E-2 & 5.62E-2 & 9.71E-2 & 1.49E-1 & 2.16E-1 & 2.96E-1 & 3.84E-1 & 4.72E-1 & 5.54E-1 & 6.26E-1 \\

& \textbf{RPLPO} & \textbf{1.41E-2} & \textbf{2.34E-2} & \textbf{3.53E-2} & \textbf{5.31E-2} & \textbf{7.80E-2} & \textbf{1.09E-1} & \textbf{1.45E-1} & \textbf{1.83E-1} & \textbf{2.20E-1} & \textbf{2.54E-1} \\
\hline
\multirow{5}{*}{NSWE} & PIDL & 2.27E-1 & 4.02E-1 & 5.40E-1 & 6.42E-1 & 7.10E-1 & 7.54E-1 & 7.86E-1 & 8.17E-1 & 8.53E-1 & 8.94E-1 \\

& PINO* & 2.24E-1 & 3.97E-1 & 5.32E-1 & 6.31E-1 & 6.97E-1 & 7.39E-1 & 7.68E-1 & 7.97E-1 & 8.32E-1 & 8.73E-1 \\

& FNO & 3.35E-1 & 3.44E-1 & 3.57E-1 & 3.74E-1 & 3.94E-1 & 4.10E-1 & 4.26E-1 & 4.53E-1 & 4.76E-1 & 4.87E-1 \\

& FNO* & 3.52E-2 & 7.34E-2 & 1.11E-1 & 1.48E-1 & 1.82E-1 & 2.12E-1 & 2.48E-1 & 2.92E-1 & 3.40E-1 & 3.91E-1 \\

& \textbf{RPLPO} & \textbf{1.64E-2} & \textbf{3.29E-2} & \textbf{5.19E-2} & \textbf{7.36E-2} & \textbf{9.49E-2} & \textbf{1.10E-1} & \textbf{1.22E-1} & \textbf{1.40E-1} & \textbf{1.63E-1} & \textbf{1.82E-1} \\
\hline
\end{tabular}
\end{sc}
\end{small}
\end{sidewaystable*}

\end{document}